%% file: main.tex
\documentclass[journal]{IEEEtran}

\usepackage[noadjust]{cite}

\newcommand*{\paperpath}{./}

\input{\paperpath/setup/preamble.tex}
\begin{document}

%
\title{Multimodal and Multiview Deep Fusion for Autonomous Marine Navigation}
%
%
%

\author{Dimitrios~Dagdilelis,
    Panagiotis~Grigoriadis,
    and~Roberto~Galeazzi
\thanks{D. Dagdilelis, Department of Electrical and Photonics Engineernig, Technical University of Denmark.}%
\thanks{P. Grigoriadis, independent contributor.}%
\thanks{R. Galeazzi, Department of Electrical and Photonics Engineernig, Technical University of Denmark.}
\thanks{Manuscript received January 1st, 2024; revised January 2nd, 2024.}}



\maketitle

\input{\paperpath/chapters/abstract}


%
\IEEEpeerreviewmaketitle

\input{\paperpath/chapters/introduction.tex}

%
%
%
%

\input{\paperpath/chapters/tables/list_of_abbreviations.tex}
\input{\paperpath/chapters/tables/list_of_symbols.tex}

\input{\paperpath/chapters/related_work.tex}

\input{\paperpath/chapters/problem_formulation.tex}
\input{\paperpath/chapters/proposed_approach.tex}

\input{\paperpath/chapters/evaluation.tex}

\input{\paperpath/chapters/conclusion}
\input{\paperpath/chapters/future_work}

\appendices

\input{\paperpath/chapters/appendix}

\section*{Acknowledgment}
This research was sponsored by the Danish Innovation Fund, The Danish Maritime Fund, the Orients Fund, and the Lauritzen Foundation through the Autonomy part of the ShippingLab project, grant number 8090-00063B. The electronic navigational charts were provided by the Danish Geodata Agency.

\ifCLASSOPTIONcaptionsoff
  \newpage
\fi

\bibliographystyle{IEEEtran}
\bibliography{references}

%







\end{document}

%% file: preamble.tex
\everymath=\expandafter{\the\everymath\displaystyle}

\newcommand{\Vset}{\mathcal{V}}

\newcommand{\Bset}{\mathcal{B}}
\usepackage{natbib}
\newcommand{\CVT}{\mathcal{X}}
\usepackage{import}
\usepackage{gensymb}

%% file: chapters/abstract.tex
\begin{abstract}
    This paper introduces a cross-attention transformer-based approach for multimodal sensor fusion, aimed at constructing a bird’s-eye view of a vessel's surrounding environment to enhance safe marine autonomous navigation. The proposed method utilizes an innovative model that performs deep fusion of multiview RGB images and long-wave infrared camera images, together with sparse LiDAR point clouds. Additionally, model training incorporates data from X-band radar and electronic nautical charts to generate relevant prediction classes. The reconstructed view offers a detailed representation of the vessel's surroundings, facilitating more accurate and robust navigation. The effectiveness of the proposed approach is evaluated using real-world data collected at sea. Experimental results demonstrate its capability across various scenarios, including challenging weather conditions and complex marine environments.
\end{abstract}

%% file: chapters/introduction.tex
\section{Introduction}\label{sec:Intro}
Highly automated and autonomous navigation has become a critical area of research in the field of marine robotics due to the growing demand for unmanned maritime operations. However, some of the key challenges in the development of safe autonomous marine systems are to enable a robust and accurate perception of the environment surrounding the vessel and to provide spatial reference for objects in such an environment.

Robust perception depends on the integration of data from diverse sensors, enabling the mitigation of individual sensor limitations through data fusion. A multisensor configuration enhances contextual awareness by leveraging the capabilities of cameras while providing robust detection capabilities—both near and far—via radar and LiDAR that can operate effectively under adverse weather and lighting conditions~\cite{Singh2023Vision-RADARSurvey}.

Accurate perception requires the determination of the 3D position of other vessels and/or navigation targets~\cite{paasche2023a}. A typical multimodal sensing platform comprises one or more cameras in conjunction with one or several sparse sensors, such as LiDAR or radar. While cameras are limited in their ability to accurately assess 3D geometries due to a lack of depth information, they excel in delivering rich semantic and boundary data. In contrast, although sparse sensors are capable of providing precise short-range 3D positional information, they often fall short in terms of dense semantic content. The integration of dense cameras with sparse sensors represents a promising avenue for advancement. However, existing research in multimodal maritime fusion, although recognizing the importance of this integration, predominantly concentrates on the narrow scope of identifying and tracking foreign vessels \cite{farahnakian2018a,ma2022a,YAO2023113939,han2019a,haghbayan2018a,douguet2023multimodal,helgesen2019a}. 

This emphasis highlights a significant gap compared to the advancements observed in \gls{AD} technology, where perception systems have achieved more comprehensive scene understanding capabilities~\cite{a2024a}. This broader understanding is crucial for supporting essential autonomous functions, such as collision avoidance and route planning, thereby facilitating effective and safe navigation in diverse environments~\cite{a2024a}.

\begin{figure*}
    \centering  
	\includegraphics[width=1.9\columnwidth]{\paperpath/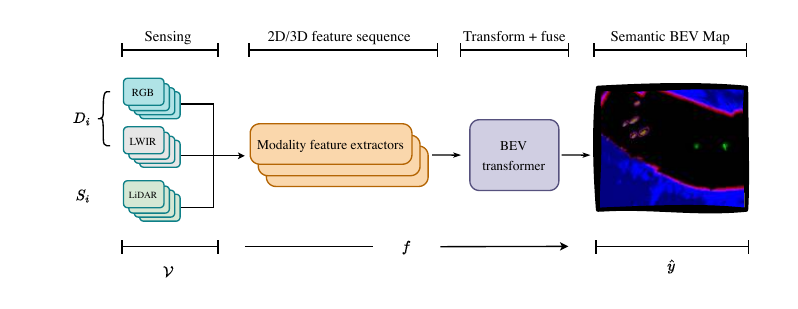}
         \caption{High level overview of the proposed approach. Sensor streams are processed by modality-specific feature extractors. Modality features are processed as a sequence by the BEV Transformer model into a semantic BEV map. The final output representation is orthographic, compact and semantically useful for downstream navigation tasks.}
	\label{fig:mother-figure}
\end{figure*}

The literature review (Sections \cref{sec:RelatedWork}-\cref{sec:Problem}) shows that no previous study has adapted or validated multimodal perception technology specifically for autonomous waterborne navigation. In light of this gap, our research proposes a perception framework designed to process multimodal sensor data from RGB and long-wave infrared (LWIR) cameras, along with LiDAR sensors. This framework aims to facilitate comprehensive scene understanding by predicting, in real-time, the \gls{BEV} positions of nearby navigation features. A high-level overview of our approach is illustrated in \cref{fig:mother-figure}.

The remainder of the paper is structured as follows. Section \cref{sec:RelatedWork} reviews the existing literature on methods for generating BEV models using multi-modal sensor platforms. Section \cref{sec:Problem} formalizes the research problem and reviews existing methods. Section \cref{sec:Approach} details the development of the proposed approach and discusses how the different sensor technologies are fused through the architecture of the cross-attention transformer. Section \cref{sec:Eval} shows the results of testing the approach using real-world data collected at sea. Sections \cref{sec:Conclusion} and \cref{sec:FutureWork} draw conclusions and discuss the current limitations of the proposed method and future work.

Table \cref{tab:abbreviations} provides the list of abbreviations used in the paper.

%% file: chapters/tables/list_of_abbreviations.tex
\begin{table}[ht]
    \centering
    \caption{Abbreviations}
    \label{tab:abbreviations}
    \begin{tabular}{m{2cm}p{.6\columnwidth}}
    \toprule
    \textbf{Abbreviation} & \textbf{Explanation} \\
    \midrule
    BEV & Bird Eye View \\
    \addlinespace
    ENC & Electronic Navigational Charts\\
    \addlinespace
    RCS & Radar Cross Section \\
    \addlinespace
    IoU & Intersection over Union\\
    \addlinespace
    XBR & X-band GHz Radar\\
    \addlinespace
    GNSS & Global Navigation Satellite System \\
    \addlinespace
    RGB & Red Green Blue \\
    \addlinespace
    LWIR & Long Wave Infrared\\
    \bottomrule
    \end{tabular}
\end{table}

%% file: chapters/tables/list_of_symbols.tex
\begin{table}[ht]
    \centering
    \caption{List of Key Mathematical Notations}\label{tab:notations}
    \begin{tabular}{m{2cm}p{.6\columnwidth}}
    \toprule
    \textbf{Symbol} & \textbf{Description} \\
    \midrule
    \( \Vset \) & Heterogeneous sensor views collection \\
    \( \Sele \) & Sparse sensor views collection \\
    \( \Dele \) & Dense sensor views collection \\
    \( \Bset \) & Multi-class binary BEV maps \\
    \( N_B \) & Number of grid coordinates in the BEV map \\
    \( C_B \) & Number of semantic BEV map classes \\
    \( V_i \) & Number of views in modality $i$ \\
    \( N_i \) & Number of elements in modality $i$ \\
    \( C_i \) & Number of channels in an element of modality $i$ \\
    \( t \) & Number of aggregated data chunk time instances \\
    \( y \) & Ground truth BEV map \\
    \( \mathcal{G}\) & Sparse modality pre-processing method \\
    \( \rawpcd\) & Raw sparse point cloud in 3D \\
    \( \pseudopcd\) & Post-processsed, dense, pseudo-camera view \\
    \( Conv2d \) & Convolutional-2D \\
    \addlinespace
    \( x^W \) & World-point in euclidean space \\
    \( x^I \) & Image pixel coordinates vector \\
    \( p_i, p_j \) & Image pixel coordinates \\
    \( d \) & Depth from image plane \\
    \( r_i \) & View-aware data feature rays \\
    \( \timeview{f} \) & Input sequence feature vector $i$, corresponding to view $k$ at time instant $\tau$ \\
    \( \voxel \) & Set of points in a voxel \\
    \( \phi(\cdot) \) & Linear layer, embedding position encodings \\
    \( \epsilon(\cdot) \) & Linear layer, embedding camera position vector \\
    \(\iota(\cdot) \) & Linear layer, embedding the temporal encodings \\
    \addlinespace
    \( q_i \) & Dense ray queries in BEV \\
    \addlinespace
    \( b_i \) & Inverse projection of pixel vectors in the context of camera parameters \\
    \addlinespace
    \( \outputsequence \) & Latent BEV map feature sequence \\
    \( Q,K,V \) & Transformer queries, keys, values \\
    \addlinespace
    \( \timeview{\rho} \) & Direction encoded feature vector \\
    \addlinespace
    \( \tilde{q}_i \) & Learnable BEV queries \\
    \( q_i \) & Position encoded BEV queries \\
    \addlinespace
    \( E_i, I_i \) & Extrinsic and Intrinsic camera matrices \\
    \( \camposition \) & Extrinsic and Intrinsic camera matrices \\
    \addlinespace
    \( H, W \) & Input view's height and width \\
    \( h_q, w_q \) & Width and height of latent BEV map representation \\
    \( n_{\text{BEV}}=h_q \times w_q \) & Number of latent BEV map tokens \\
    \( \mathcal{U} \) & BEV map segmentation decoder \\
    
    \bottomrule
    \end{tabular}
\end{table}

%% file: chapters/related_work.tex
\section{Related Work}\label{sec:RelatedWork}

\textbf{Bird's Eye View:}
\gls{BEV} recognition models \cite{li2022bevformer,liu2022petr,Zhou2022Cross-viewSegmentation,Pan2019Cross-viewSurroundings,Roddick2018OrthographicDetection} represent a category of models employed for 3D object detection. These models have garnered both the industry's and academia's research attention, due to their ability to merge partial raw data from heterogeneous perception sensors into a cohesive and comprehensive 3D output space.

Combining a dense camera with sparse sensors is a promising direction; however, the data collected by these sensors must be mapped to a unified coordinate frame. Naturally, \gls{BEV} emerges as the ideal representation framework to support such uniform coordinate mapping. More specifically, a \gls{BEV} representation:

\begin{enumerate*}[label=\alph*)]
    \item is independent of the sensor model and thus easily extendable to additional modalities,
    \item is geometrically interpretable,
    \item is the framework that is common to all perception modalities, but at the same time native to downstream navigation tasks, such as motion planning.
\end{enumerate*}

According to \cite{li_delving_2023}, existing BEV fusion methods are limited because of two reasons. Despite the fact that sparse modalities like LiDAR, can be trivially used to predict 3D features, the same does not apply for camera modalities, where depth information is implicit or missing, and extracting 3D spatial information from monocular or multi-view settings is difficult and largely unsolved~\cite{li_delving_2023}. At the same time, many sensor fusion algorithms rely on simple object-level fusion or basic feature concatenation, where misalignment or inaccurate depth predictions between camera and LiDAR data can lead to sub-par performance compared to LiDAR-only methods~\cite{li_delving_2023}.

To overcome these limitations, deep fusion methods have emerged as a promising approach that leverages deep learning models to integrate data from multiple sensors~\cite{li_delving_2023}. Among the deep fusion methods, the transformer-based model has shown superior performance in capturing the complex relationships between different modalities and fusing them into a comprehensive representation of the surrounding environment~\cite{li_delving_2023}. 

In this paper, we propose a novel cross-attention transformer-based multimodal sensor fusion approach for marine autonomous navigation. The proposed method integrates information from multiple sensors, including LiDAR, RGB, LWIR, ENC charts, and GPS, and leverages a deep learning model to effectively capture complex relationships between different modalities, in order to predict a semantic \gls{BEV} map of the surrounding environment. Semantic \gls{BEV} maps can be used to enable downstream navigation tasks, or be used to perform diagnostics on other sensor modalities such as the maritime radar.

Our experimental results demonstrate the effectiveness of the proposed method in coastal marine navigation environments. Finally, our work contributes to the development of more accurate and robust marine autonomous systems by proposing a deep fusion approach that
\begin{enumerate*}[label=\alph*)]
\item combines sparse and dense perception data,
\item integrates the temporal information,
\item remains robust to sensor-calibration errors.
\end{enumerate*}

%% file: chapters/problem_formulation.tex
\section{Problem formulation}\label{sec:Problem}


The core idea of BEV perception is to develop a feature representation of information, directed from possibly multiple views or modalities, that can be readily used to solve a downstream task, such as 3D object detection, BEV map segmentation, lane line detection, path planning, or any combination of these tasks. In the following sections, we use \emph{2D} to refer to a perspective view described by camera pixel coordinates, \emph{3D} to refer to the real-world space with world coordinates, and \emph{BEV} to refer to a planar, metric, zero height, own-ship centered, \emph{Bird’s Eye View} world  coordinate grid (see \cref{fig:target-extraction,fig:top_down_example_real}). 

\subsection{Task definition}
The task on which we focus is that of \emph{BEV map segmentation}, i.e., given the features of \textit{ multiple} dense or sparse modalities, we are tasked with predicting a \textit{ orthographic semantic} BEV map; the task can be formally described as:

Let a heterogeneous collection of multi-modal features $\Vset$, where

\begin{equation}
\Vset = \{\Dele_1, \ldots, \Dele_i\} \cup \{\Sele_1, \ldots, \Sele_j\}
\label{eq:heterogeneous-data}
\end{equation}

and
\begin{description}[left=15pt]
    \item[$\Dele_i$] $\in \reals^{t \times V_i \times N_i \times C_i}$ and $i$ is a dense 2D modality index.
    
    \item[$\Sele_j$] $\in \reals^{t \times V_j \times N_j \times C_i}$ and $j$ is a sparse 3D modality index.
    
    \item[$t, V_k, N_k, C_k$] Represent the number of aggregated time instances, views, elements, and channels of each modality $k$ respectively, with $k \in \{i, j\}$.
\end{description}

Let also a binary multi-class BEV map $y \in \Bset $, where

\[ \Bset \subseteq \{0,1\}^{N_B \times C_B} \]

with $N_B, C_B$ respectively being the number of elements (map resolution) and classes of the BEV map (see \cref{fig:target-features}).

BEV map segmentation, finds a function $f$ that transforms heterogeneous views \(\Vset\) to binary BEV masks \( \hat{y}\), i.e.
 \[\hat{y} = f(\mathbf{\Vset}) \quad \text{where} \quad \hat{y} \in \Bset \]

This problem is particularly complex due to the fact that its inputs and outputs operate within distinct coordinate systems. Structured inputs $D_i$ are captured from calibrated camera views in 2D perspective view, point cloud inputs $S_i$ are captured in 3D, and outputs $y$ are predicted in BEV. 

\subsection{Existing methodologies}

Existing approaches that model $f$ are characterized depending on:
\begin{enumerate*}[label=\alph*)]
\item \textit{supported modalities},
\item \textit{modality feature extraction mechanics},
\item \textit{2D feature transformation module}.
\item \textit{feature fusion mechanics}, 
\end{enumerate*}

\textbf{Supported modalities:}
Camera-only 3D perception is a significant focus in academic research \cite{wang2021fcos3d,liu2020smoke,huang2021bevdet,liu2022petr,li2022bevformer,liu2023bevfusion,reiher2020sim2real,hu2021fiery,li2022hdmapnet,philion2020lift,hu2022st} since it avoids the need of using expensive lidar equipment. The fundamental challenge in camera-only 3D perception, lies in the fact that 2D imaging processes inherently lacks the ability to capture 3D information, making accurate object localization difficult without precise depth estimation. 

Lidar only approaches \cite{shi2020pv, yin2021center, fan2022embracing, hu2022afdetv2, mao2021pyramid, chen2022mppnet, zhou2018voxelnet, lang2019pointpillars}, completely forgo the use of cameras, and hence suffer performing in dense downstream prediction tasks. On the other hand, approaches similar to ours, combine the best of both worlds by fusing camera with lidar features \cite{vora2020pointpainting, yin2021multimodal, chen2022autoalign, li2022deepfusion, Liu2022BEVFusion:Representation}.

\textbf{Feature extraction:}
The most widespread approach in multi-modal feature extraction is having individual encoders extracting features from each modality. Concerning image feature extraction, there exists a wealth of 2D perception research, in the form of transfer learning and pre-trained feature extraction backbones \cite{he2019rethinking}. In lidar-only approaches, \cite{zhou2018voxelnet, shi2020pv, fan2022embracing, yan2018second, he2020structure,mao2021voxel} use voxelization to structure point-clouds into voxels, followed by a 3D feature extractor, thus retaining data 3D structure but increasing computations. Similarly to our approach, \cite{lang2019pointpillars, Chen2016Multi-ViewDriving,yang2018pixor,li2022hdmapnet,yang2018hdnet,zeng2018rt3d,ali2018yolo3d,simony2018complex} convert point cloud data into a BEV representation, by discretizing points into a BEV grid. Features such as height, intensity, and density are then extracted from the points within each grid cell to represent the grid's features. However, due to the large number of points in each BEV grid cell, this process can lead to significant information loss.

\textbf{Feature fusion:} Depending on the stage of information fusion we can classify the approaches in two types:
\begin{enumerate}
    \item Early fusion, refers to combining information at an early stage of processing, such as combining \emph{raw sensor inputs}.
    \item Deep fusion, facilitates the interaction of latent modality features within a neural network's structure, leveraging their flexibility to expressing non-linear functions between modalities.
\end{enumerate}

Early fusion in BEV perception operates by decorating one modality with features from the others. In doing so, there is no single optimal decision on which modality is going to be the main carrier of information, as all selections have drawbacks.  A widely used early-fusion technique is Painting \cite{Wang2021CVPR,vora2020pointpainting}. Painting projects point cloud data to images to create correspondences and subsequently appends semantic information to the points, while discarding the rest of the information in the images.  Point-level decoration with image features is semantically lossy because it suffers from throwing away a lot of contextual information from the cameras due to point-cloud sparsity, which has a severe impact on semantic-oriented tasks like BEV segmentation, while the reciprocal process of image feature decoration is geometrically lossy due to perspective view geometry \cite{Liu2022BEVFusion:Representation}.

Deep fusion leverages the network's capacity to learn complex representations, allowing non-linear interaction between modalities. \cite{liu2023bevfusion,liang2022bevfusionsimplerobustlidarcamera} propose an effective fusion method to transform 2D camera features to BEV features, by efficiently projecting camera features into BEV space and then combining them with lidar BEV features using convolutional layers. In a similar approach, \cite{liang2022bevfusionsimplerobustlidarcamera,li2022unifyingvoxelbasedrepresentationtransformer} used predicted image depth distributions and common 3D convolutions to generate modality specific voxel spaces that communicate during task prediction, enhancing cross-modal interactions.

\textbf{Feature transformation}
Point clouds bear 3D geometry information, and can therefore be transformed in BEV by geometry projection. Camera views on the other hand, lack depth information and transforming them to BEV is non trivial. Existing multi-modal feature fusion methods, are highly dependent on a reliable 2D to BEV image feature transformation. Methodologies for transforming 2D features into BEV representations can be classified into three primary categories based on their approach to depth estimation:
\begin{enumerate}
    \item Those employing explicit depth prediction.
    \item Those utilizing implicit depth prediction.
    \item Those that operate without any depth prediction.
\end{enumerate}
This categorization reflects the varying strategies used to address the challenge of projecting planar image features into a three-dimensional space. 

\begin{figure}
    \centering  
	\includegraphics[width=0.9\columnwidth]{\paperpath/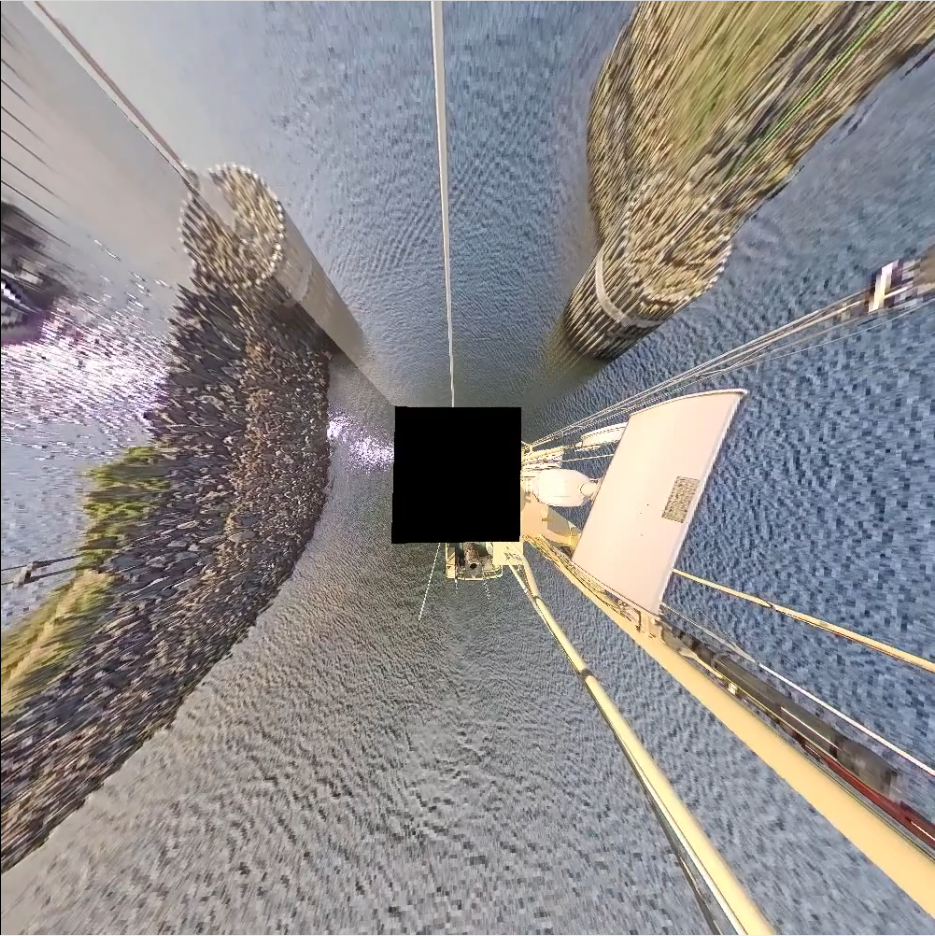}
         \caption{Outlook from a virtual camera in a top down view pose, based on \SI{360}{\degree} images from four cameras. The scene is the entrance to a leisure craft harbor at Limfjorden (DK) \cref{MartinsPaper}}
	\label{fig:top_down_example_real}
\end{figure}

Without performing depth prediction, Inverse Perspective Mapping (IPM) \cite{mallot1991inverse} introduced a homography derived from the camera's int/extrinsic parameters, and projects from 2D to 3D and vice versa, assuming that the corresponding 3D points lie on a horizontal plane. The basic idea has been used in recent work  \cite{Garnett3D-LaneNet:Detection, Kim2019DeepImage, LoukkalDrivingPlanning, Philion2020Lift3D}, with \cite{MartinsPaper,paasche2023a} applying for free-space estimation. Violation of the planar assumption  and calibration noise create strong artifacts, while performance degrades quickly as distances get longer due lower pixel density at vanishing points and projective geometry~(\cref{fig:top_down_example_real}).

Lift-splat-shoot (LSS) \cite{philion2020lift}, is a pioneering approach that uses a pre-trained monocular depth prediction model to predict the depth distribution of image features and use it to project them in 3D, addressing the challenging camera-to-BEV transformation problem. Their method resembles pseudo-lidar \cite{DBLP:journals/corr/abs-1812-07179}, where depth is used to lift dense pixels into 3d points.  Subsequent work in \cite{DBLP:journals/corr/abs-2103-01100} employs a similar approach to LSS to predict categorical depth distribution, but unlike LSS that uses pre-trained depth models, it provides supervision for depth prediction. This line of work is followed by \cite{huang2021bevdet, li2023bevdepth, liu2023bevfusion,guo2021a,Park2021detection} and extended to stereo vision depth prediction by \cite{guo2021a,chen2020a}.

Several BEV perception works utilize either multi-layer perceptron \cite{li2022hdmapnet,saha2022translating,pan2020cross} or the transformer architecture \cite{vaswani2017attention}, and implicitly perform depth prediction, in order to transform 2D features to BEV. In a similar way, CVT \cite{Zhou2022Cross-viewSegmentation}, which greatly influences our work, transforms 2D features to BEV by implicitly predicting depth, and uses learnable camera aware embeddings and the cross attention mechanism to query modality features from BEV coordinates and construct BEV features.

\textbf{Calibration:}
Most of the previous work in early and late fusion approaches requires knowledge of precise extrinsic and intrinsic sensor calibration parameters to fuse features between the different coordinate frames, perspectives, and fields of view \cite{MartinsPaper}. Unfortunately, the calibration process can be inaccurate, as illustrated in \cref{fig:calib-errors}, where world coordinates are projected on image pixels using camera geometry and calibration parameters. The convolution blocks utilized in most existing approaches \cite{Liu2022BEVFusion:Representation, Harley2022Simple-BEV:Perception, Li2021HDMapNet:Framework} cannot compensate for dynamic calibration errors, due to their translation invariance. The multiplicity of the sensors, the laborious, time-consuming, and expensive process of calibrating a multi-sensor platform, and the fact that after calibration, the system's performance remains sensitive to external perturbations, have created the need for calibration-free fusion methods. To this end, researchers have explored implicit parameter learning in their models \cite{Liu2022PETR:Detection, Zhou2022Cross-viewSegmentation}, similar to our approach.

\subsection{Key problems}

BEV perception aims to develop a resilient and adaptable feature representation using both camera and LiDAR data. This process is straightforward for LiDAR input, as point clouds inherently contain 3D information. However, it's more challenging for camera input, where extracting 3D spatial data from single or multiple 2D views poses significant difficulties.

Another crucial challenge lies in effectively merging features during the early or middle stages of the processing pipeline. Many existing sensor fusion methods oversimplify this process, either by combining objects at a high level or by naively concatenating features along the data channel. This approach often leads to sub-optimal performance, with some fusion algorithms actually under-performing compared to LiDAR-only solution \cite{li_delving_2023}. The poor performance can be attributed to misalignment issues or inaccurate depth estimates when integrating camera and LiDAR data. Consequently, developing methods to properly align and integrate features from multiple input modalities is critical and presents significant opportunities for innovation in this field.

%% file: chapters/proposed_approach.tex
\section{Proposed Approach}\label{sec:Approach}

Our proposed method, extends the camera-only CVT \cite{Zhou2022Cross-viewSegmentation} by \begin{enumerate*}[label=\alph*)]
\item appending lidar and LWIR cameras to the operating modalities, 
\item considering point clouds as pseudo-images and pseudo-views, and processing them in parallel with camera views, 
\item integrating temporal information from previous timesteps in the fusion process.
\end{enumerate*}

To do so, we extract features from individual sensor streams, using modality-specific encoders. Thereafter, we utilize a view transformation module,  that utilizes cross-attention guided by position-aware camera encodings, as well as learnable BEV queries, to transform multi-modal sensor features to BEV features. Finally, a decoder upscales the BEV features to the original ground truth BEV map resolution. We provide a high level overview of our approach in \cref{fig:mother-figure}, while in \cref{fig:cvt} we provide a more analytical description of the model's components. The model in \cref{fig:cvt} is end-to-end differentiable, and we optimize it using ground truth BEV maps (read \cref{subseq:dataset}) and focal-loss \cite{lin2017focal}.

\input{\paperpath/chapters/proposed_approach/dataset}

\input{\paperpath/chapters/proposed_approach/solution_architecture}
\input{\paperpath/chapters/proposed_approach/multimodality}
\input{\paperpath/chapters/proposed_approach/temporality}

%% file: chapters/proposed_approach/dataset.tex
\subsection{Dataset}\label{subseq:dataset}

\begin{table*}[htbp]
\caption{Sensor Specifications}
\label{tab:sensor_specifications}
\centering
\resizebox{\textwidth}{!}{%
\begin{tabular}{l c c c c c c c}
\toprule
\textbf{Sensor} & \textbf{Max Range} & \textbf{Min Range} & \textbf{Ang. Res.} & \textbf{Range Res.} & \textbf{FoV} & \textbf{Effective FOV} \\
\midrule
X-band radar & 20 NM & 85 M & 0.225 degrees & 1\% Max range & 360 deg & 360 \\
RGB cam & -- & -- & 0.0477 deg & -- &  (94, 52) deg & 360 deg \\
LWIR cam & -- & -- & 0.078 deg & -- &  (50, 40) deg & 200 deg \\
Lidar & 200m & 0.3 m & (0.18, 0.7) deg & 12 cm & (360, 45) deg & 200 deg \\
ENC & -- & -- & -- & -- & --  & 360 deg \\
\bottomrule
\end{tabular}
}
\end{table*}

\begin{figure}
    \centering  
	\includegraphics[trim=200pt 360pt 100pt 220pt, clip,  width=0.8\columnwidth]{\paperpath/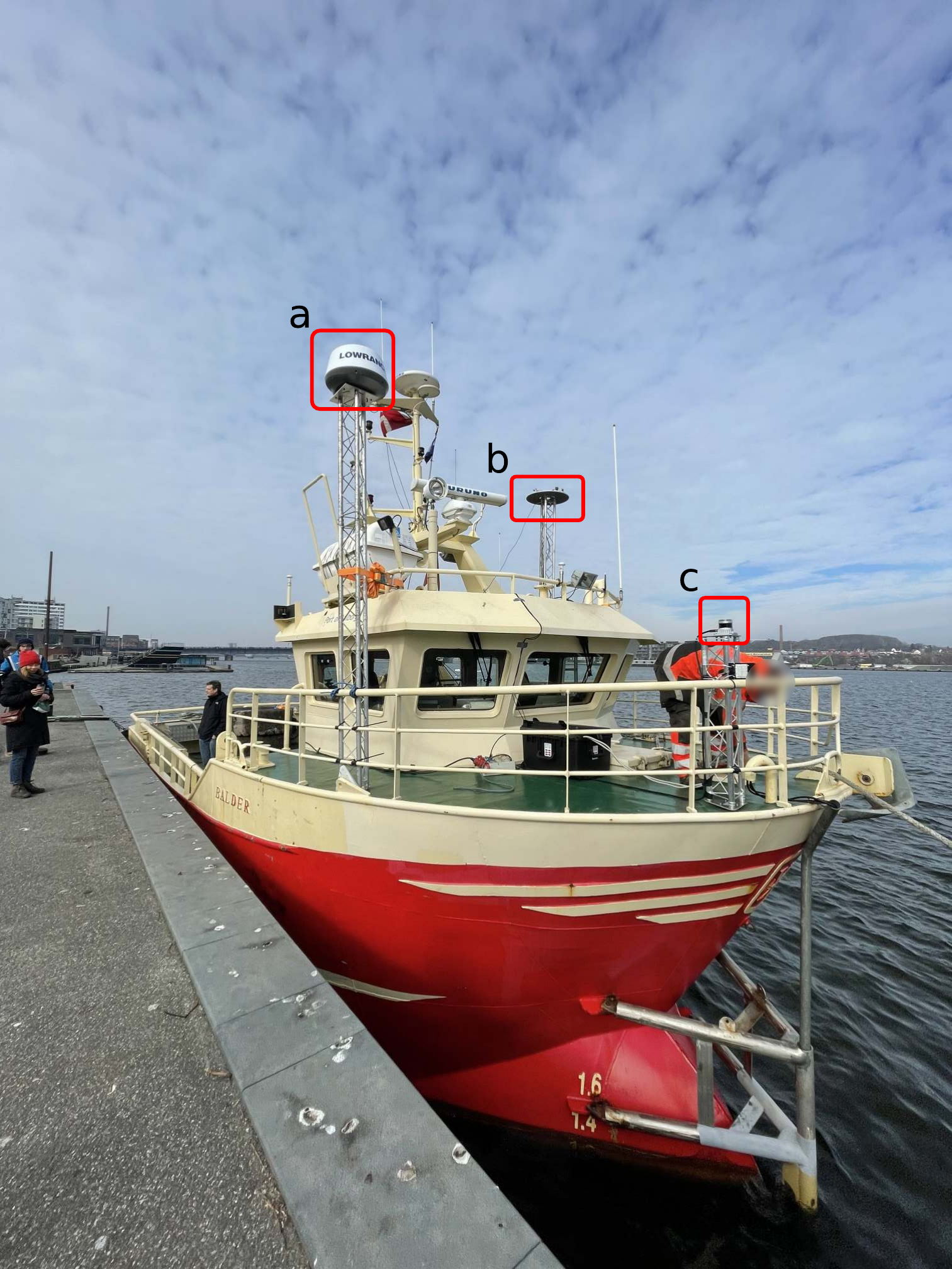}
         \caption{Tugboat \emph{Balder} used for the data collection at \textit{Limfjorden, DK}. Annotated are: a) XBR b) RGB \& LWIR camera platform c) LiDAR}
	\label{fig:balder_sensor_stack}
\end{figure}

Our maritime specific dataset, was collected in Aalborg, Denmark, onboard tugboat \emph{Balder}, during daytime and over a duration of 6 hours. The dataset includes diverse set of modalities, including RGB, LWIR images, two lidar sensors positioned at the front and rear of the vessel and an X-band long-range radar. See \cref{tab:sensor_specifications} for a summary of sensor specifications. Additionally, the dataset includes own-ship's geo-location, captured from a satellite compass and a GNSS receiver and static ENC data. Sensors were rigidly mounted on 3m masts (see \cref{fig:balder_sensor_stack}), in order to establish a prominent position and minimize obstructions from vessel structures or other sensors.

\begin{figure}
    \centering  
	\includegraphics[width=0.9\columnwidth]{\paperpath/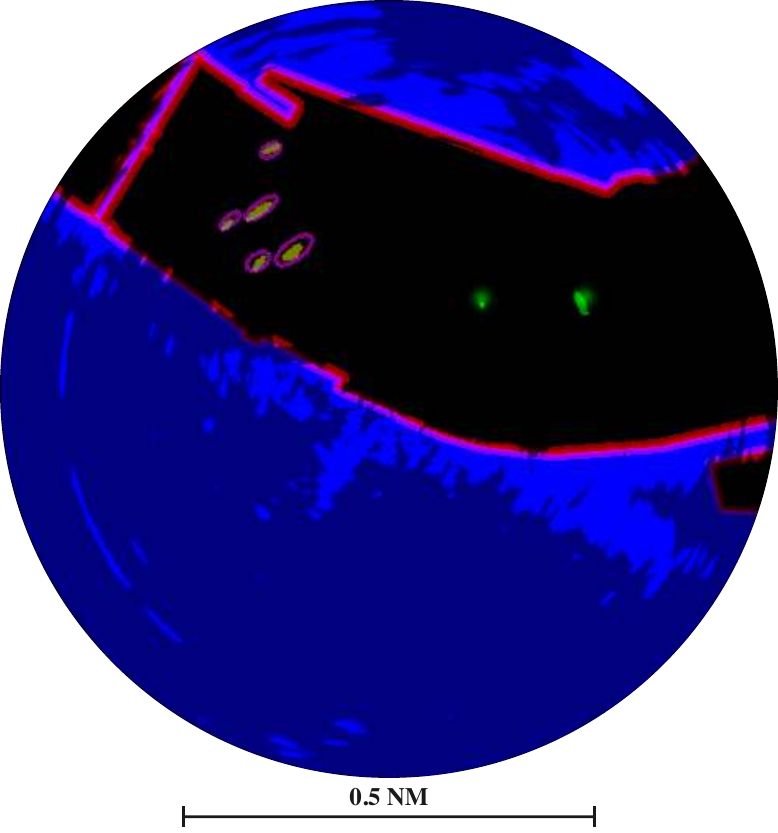}
         \caption{Ground truth BEV map generation. XBR data are transparently ploted below color-coded ENC classes (blue for land, magenta for shoreline, green for buoy, black for water). We curate \emph{moving target} instances, by annotating  magenta ellipsoids on top of verified moving targets.}
	\label{fig:target-extraction}
\end{figure}

We exploit the complementary nature of the available data sources to construct multi-class BEV ground truth maps \(y \in \Bset \) as illustrated in \cref{fig:target-extraction}.

ENC are integral to navigation systems and software used on ships and various marine vehicles, playing a critical role in facilitating safe sea travel. The information in them is gathered through surveys and observations conducted by maritime authorities. The information that is relevant in the context of our application, is the location of buoys, land, sail-able water area, and shorelines. We pre-process this information by fusing it with other sensor modalities, generating supervision signals for our model, in the form of BEV maps. To do so, we fuse ENC, with XBR and own-ship geo-location and attitude data, to generate BEV ground truth maps that semantically encode floating buoys, land, shoreline, water volumes and moving targets. 

\textbf{Static classes:}
The position of objects such as navigational marks, man-made structures, and landscape features, does not change during navigation. The geo-location of such features in the navigation scene, can be trivially extracted from the ENC, requiring no human annotation effort. Our solution aims to take advantage of the following static ENC features:

\begin{enumerate}
    \item Buoys: Charts detail the precise locations and characteristics of navigational aids, including buoys. Specifically, they provide information about the buoy's type, color, shape, and light characteristics, to assist mariners in identifying and navigating around them.
    \item Land areas: Areas occupied by landscape or structures.
    \item Shorelines: Outline shape of land areas.
    \item Sail-able area: We threshold depth-map contours to extract sail-able water area.
\end{enumerate}

\textbf{Moving targets:}
Floating objects within the navigable area are likely to be detected by the XBR, but they may not appear on the ENC. To determine the geographic location of these targets, we use the own-ship's geo-location and attitude measurements to project the XBR data onto navigable water area maps, creating XBR-over-water maps. Since XBR data over water is not affected by ground clutter, it exhibits a high signal-to-noise ratio, making it probable that echoes detected in these XBR-over-water maps correspond to actual moving targets. We then manually review the identified blobs in the XBR-over-water maps, cross-referencing them with camera data (see \cref{fig:calib-errors}) to confirm the presence of moving targets and eliminate false positives. This process allows us to accurately determine the geographic location of moving targets.

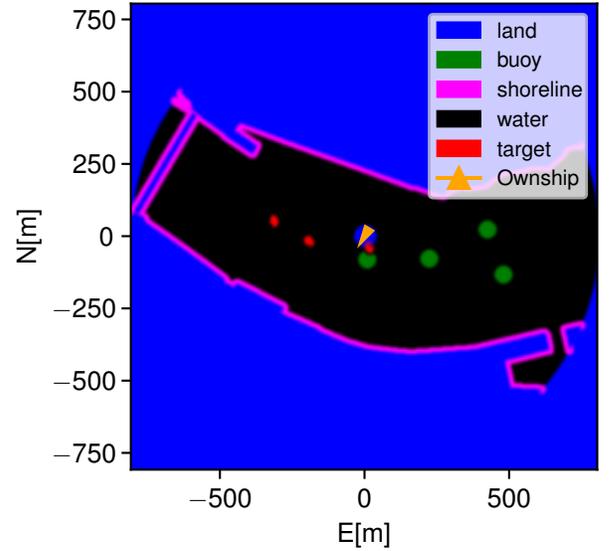
\begin{figure}[ht]
    \centering
    \import{\paperpath/figures/tex/bev-targets-raw}{bev-targets-raw_140_time1657876089.234451.tex}
    \caption{Creating a ground truth BEV map $y$ using XBR, ENC, satellite-compass, and GNSS. }
    \label{fig:target-features}
\end{figure}

\begin{figure*}[ht]
    \centering
    \resizebox{1.0\linewidth}{!}{
    \includegraphics[]{\paperpath/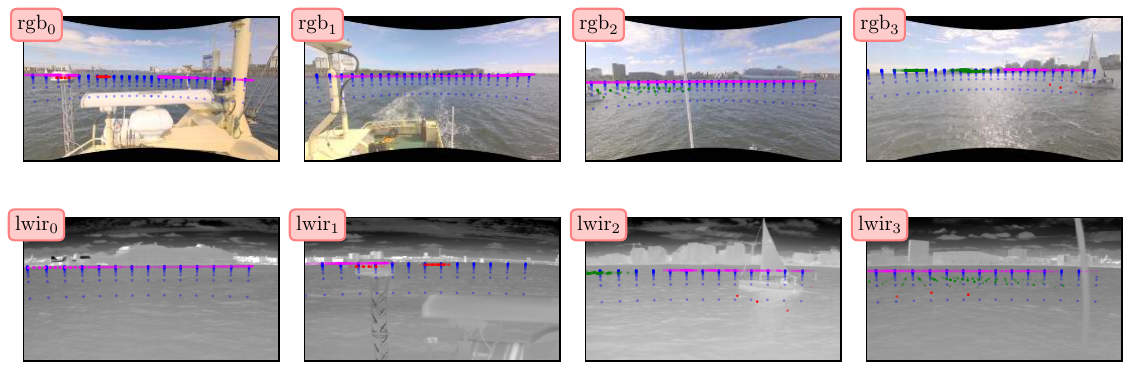}
    }
    \caption{Camera view corresponding to \cref{fig:target-features}. The semantic map $y$ in \cref{fig:target-features} is sparsely projected on the camera views (note the color inversion between blue and black). Projection errors are visible in camera views, due to inaccurate calibration parameters or noise in the own-ship pose.}
    \label{fig:calib-errors}
\end{figure*}

\begin{figure}[ht]
    \centering
    \resizebox{\columnwidth}{!}{
    \import{\paperpath/figures/tex/prediction-heatmaps}{prediction-heatmaps_140_time1657876089.234451.tex}
    }
    \caption{Softmaxed BEV map predictions $\hat{y}$. Each BEV map in the figure's tiles, predicts the presence of one of the 5 classes of interest, within 600x600m area around own-ship, where pixel intensity encodes the class's likelihood.}
    \label{fig:prediction-heatmap}
\end{figure}
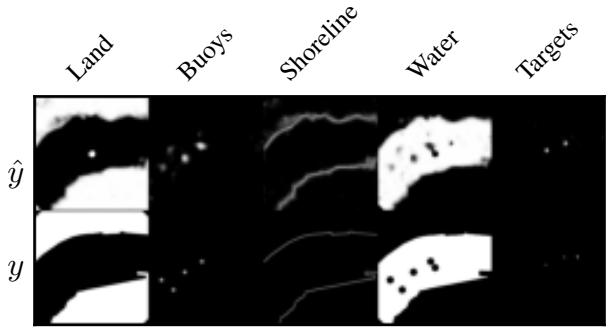

\begin{figure}[ht]
    \centering
    \resizebox{1.0\linewidth}{!}{%
    \import{\paperpath/figures/tex/targets-BEV-bodyframe-cameras-FOV}{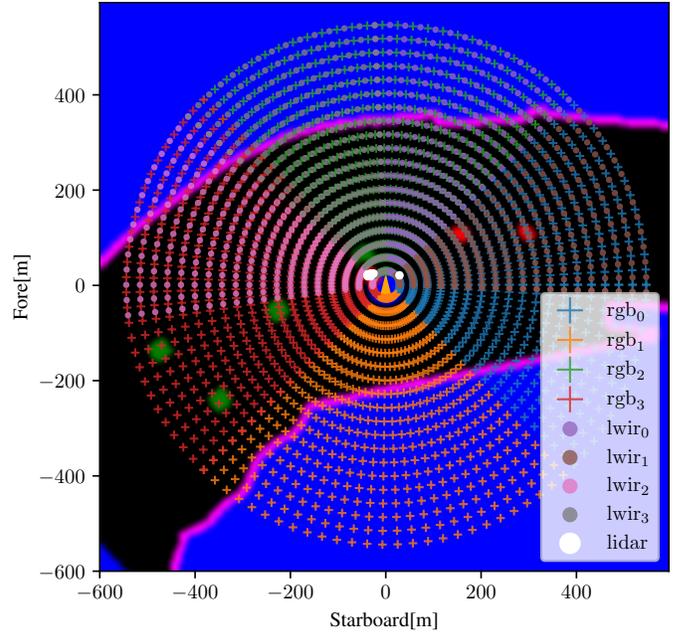}
     }
    \caption{Individual modality field of views annotated by maker type and color, projected in BEV, illustrating field of view overlap.}
    \label{fig:multi-cam-fov}
\end{figure}

 High-frequency features and XBR noise are filtered by applying a series of dilation erosion and Gaussian smoothing. The derived XBR-over-water maps, are manually annotated, using camera images as cross-referencing data, marking the BEV position of the \emph{targets} class (see \cref{fig:target-extraction}) with ellipsoids.
 
  The orthographic property of BEV maps, enables a direct and consistent relationship between pixel coordinates in the BEV map and local North-East-Down coordinates. While not strictly correct, the assumption of all BEV features residing on the same plane is valid in our operation environment. Consequently, the position of any instance or object within a BEV map can be converted to geodetic coordinates (latitude, longitude, and altitude) through the application of a local geodetic model. 
  The process involves an initial conversion from pixel coordinates to local Cartesian coordinates, followed by a transformation to geodetic coordinates using the specified local geodetic framework.

%% file: figures/tex/bev-targets-raw/bev-targets-raw_140_time1657876089.234451.tex
\begingroup%
\makeatletter%
\begin{pgfpicture}%
\pgfpathrectangle{\pgfpointorigin}{\pgfqpoint{3.249276in}{3.056373in}}%
\pgfusepath{use as bounding box, clip}%
\begin{pgfscope}%
\pgfsetbuttcap%
\pgfsetmiterjoin%
\definecolor{currentfill}{rgb}{1.000000,1.000000,1.000000}%
\pgfsetfillcolor{currentfill}%
\pgfsetlinewidth{0.000000pt}%
\definecolor{currentstroke}{rgb}{1.000000,1.000000,1.000000}%
\pgfsetstrokecolor{currentstroke}%
\pgfsetdash{}{0pt}%
\pgfpathmoveto{\pgfqpoint{0.000000in}{0.000000in}}%
\pgfpathlineto{\pgfqpoint{3.249276in}{0.000000in}}%
\pgfpathlineto{\pgfqpoint{3.249276in}{3.056373in}}%
\pgfpathlineto{\pgfqpoint{0.000000in}{3.056373in}}%
\pgfpathlineto{\pgfqpoint{0.000000in}{0.000000in}}%
\pgfpathclose%
\pgfusepath{fill}%
\end{pgfscope}%
\begin{pgfscope}%
\pgfsetbuttcap%
\pgfsetmiterjoin%
\definecolor{currentfill}{rgb}{1.000000,1.000000,1.000000}%
\pgfsetfillcolor{currentfill}%
\pgfsetlinewidth{0.000000pt}%
\definecolor{currentstroke}{rgb}{0.000000,0.000000,0.000000}%
\pgfsetstrokecolor{currentstroke}%
\pgfsetstrokeopacity{0.000000}%
\pgfsetdash{}{0pt}%
\pgfpathmoveto{\pgfqpoint{0.708026in}{0.515123in}}%
\pgfpathlineto{\pgfqpoint{3.149276in}{0.515123in}}%
\pgfpathlineto{\pgfqpoint{3.149276in}{2.956373in}}%
\pgfpathlineto{\pgfqpoint{0.708026in}{2.956373in}}%
\pgfpathlineto{\pgfqpoint{0.708026in}{0.515123in}}%
\pgfpathclose%
\pgfusepath{fill}%
\end{pgfscope}%
\begin{pgfscope}%
\pgfpathrectangle{\pgfqpoint{0.708026in}{0.515123in}}{\pgfqpoint{2.441250in}{2.441250in}}%
\pgfusepath{clip}%
\pgfsys@transformshift{0.708026in}{0.515123in}%
\pgftext[left,bottom]{\includegraphics[interpolate=true,width=2.450000in,height=2.450000in]{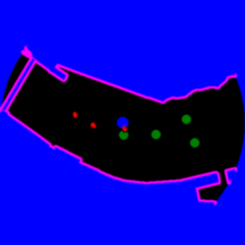}}%
\end{pgfscope}%
\begin{pgfscope}%
\pgfpathrectangle{\pgfqpoint{0.708026in}{0.515123in}}{\pgfqpoint{2.441250in}{2.441250in}}%
\pgfusepath{clip}%
\pgfsetbuttcap%
\pgfsetmiterjoin%
\definecolor{currentfill}{rgb}{1.000000,0.647059,0.000000}%
\pgfsetfillcolor{currentfill}%
\pgfsetlinewidth{1.003750pt}%
\definecolor{currentstroke}{rgb}{1.000000,0.647059,0.000000}%
\pgfsetstrokecolor{currentstroke}%
\pgfsetdash{}{0pt}%
\pgfpathmoveto{\pgfqpoint{1.907045in}{1.698007in}}%
\pgfpathlineto{\pgfqpoint{1.975242in}{1.766356in}}%
\pgfpathlineto{\pgfqpoint{1.935055in}{1.790408in}}%
\pgfpathlineto{\pgfqpoint{1.907045in}{1.698007in}}%
\pgfpathclose%
\pgfusepath{stroke,fill}%
\end{pgfscope}%
\begin{pgfscope}%
\pgfsetbuttcap%
\pgfsetroundjoin%
\definecolor{currentfill}{rgb}{0.000000,0.000000,0.000000}%
\pgfsetfillcolor{currentfill}%
\pgfsetlinewidth{0.803000pt}%
\definecolor{currentstroke}{rgb}{0.000000,0.000000,0.000000}%
\pgfsetstrokecolor{currentstroke}%
\pgfsetdash{}{0pt}%
\pgfsys@defobject{currentmarker}{\pgfqpoint{0.000000in}{-0.048611in}}{\pgfqpoint{0.000000in}{0.000000in}}{%
\pgfpathmoveto{\pgfqpoint{0.000000in}{0.000000in}}%
\pgfpathlineto{\pgfqpoint{0.000000in}{-0.048611in}}%
\pgfusepath{stroke,fill}%
}%
\begin{pgfscope}%
\pgfsys@transformshift{1.174246in}{0.515123in}%
\pgfsys@useobject{currentmarker}{}%
\end{pgfscope}%
\end{pgfscope}%
\begin{pgfscope}%
\definecolor{textcolor}{rgb}{0.000000,0.000000,0.000000}%
\pgfsetstrokecolor{textcolor}%
\pgfsetfillcolor{textcolor}%
\pgftext[x=1.174246in,y=0.417901in,,top]{\color{textcolor}{\sffamily\fontsize{10.000000}{12.000000}\selectfont\catcode`\^=\active\def^{\ifmmode\sp\else\^{}\fi}\catcode`\%=\active\def
\end{pgfscope}%
\begin{pgfscope}%
\pgfsetbuttcap%
\pgfsetroundjoin%
\definecolor{currentfill}{rgb}{0.000000,0.000000,0.000000}%
\pgfsetfillcolor{currentfill}%
\pgfsetlinewidth{0.803000pt}%
\definecolor{currentstroke}{rgb}{0.000000,0.000000,0.000000}%
\pgfsetstrokecolor{currentstroke}%
\pgfsetdash{}{0pt}%
\pgfsys@defobject{currentmarker}{\pgfqpoint{0.000000in}{-0.048611in}}{\pgfqpoint{0.000000in}{0.000000in}}{%
\pgfpathmoveto{\pgfqpoint{0.000000in}{0.000000in}}%
\pgfpathlineto{\pgfqpoint{0.000000in}{-0.048611in}}%
\pgfusepath{stroke,fill}%
}%
\begin{pgfscope}%
\pgfsys@transformshift{1.931097in}{0.515123in}%
\pgfsys@useobject{currentmarker}{}%
\end{pgfscope}%
\end{pgfscope}%
\begin{pgfscope}%
\definecolor{textcolor}{rgb}{0.000000,0.000000,0.000000}%
\pgfsetstrokecolor{textcolor}%
\pgfsetfillcolor{textcolor}%
\pgftext[x=1.931097in,y=0.417901in,,top]{\color{textcolor}{\sffamily\fontsize{10.000000}{12.000000}\selectfont\catcode`\^=\active\def^{\ifmmode\sp\else\^{}\fi}\catcode`\%=\active\def
\end{pgfscope}%
\begin{pgfscope}%
\pgfsetbuttcap%
\pgfsetroundjoin%
\definecolor{currentfill}{rgb}{0.000000,0.000000,0.000000}%
\pgfsetfillcolor{currentfill}%
\pgfsetlinewidth{0.803000pt}%
\definecolor{currentstroke}{rgb}{0.000000,0.000000,0.000000}%
\pgfsetstrokecolor{currentstroke}%
\pgfsetdash{}{0pt}%
\pgfsys@defobject{currentmarker}{\pgfqpoint{0.000000in}{-0.048611in}}{\pgfqpoint{0.000000in}{0.000000in}}{%
\pgfpathmoveto{\pgfqpoint{0.000000in}{0.000000in}}%
\pgfpathlineto{\pgfqpoint{0.000000in}{-0.048611in}}%
\pgfusepath{stroke,fill}%
}%
\begin{pgfscope}%
\pgfsys@transformshift{2.687948in}{0.515123in}%
\pgfsys@useobject{currentmarker}{}%
\end{pgfscope}%
\end{pgfscope}%
\begin{pgfscope}%
\definecolor{textcolor}{rgb}{0.000000,0.000000,0.000000}%
\pgfsetstrokecolor{textcolor}%
\pgfsetfillcolor{textcolor}%
\pgftext[x=2.687948in,y=0.417901in,,top]{\color{textcolor}{\sffamily\fontsize{10.000000}{12.000000}\selectfont\catcode`\^=\active\def^{\ifmmode\sp\else\^{}\fi}\catcode`\%=\active\def
\end{pgfscope}%
\begin{pgfscope}%
\definecolor{textcolor}{rgb}{0.000000,0.000000,0.000000}%
\pgfsetstrokecolor{textcolor}%
\pgfsetfillcolor{textcolor}%
\pgftext[x=1.928651in,y=0.238889in,,top]{\color{textcolor}{\sffamily\fontsize{10.000000}{12.000000}\selectfont\catcode`\^=\active\def^{\ifmmode\sp\else\^{}\fi}\catcode`\%=\active\def
\end{pgfscope}%
\begin{pgfscope}%
\pgfsetbuttcap%
\pgfsetroundjoin%
\definecolor{currentfill}{rgb}{0.000000,0.000000,0.000000}%
\pgfsetfillcolor{currentfill}%
\pgfsetlinewidth{0.803000pt}%
\definecolor{currentstroke}{rgb}{0.000000,0.000000,0.000000}%
\pgfsetstrokecolor{currentstroke}%
\pgfsetdash{}{0pt}%
\pgfsys@defobject{currentmarker}{\pgfqpoint{-0.048611in}{0.000000in}}{\pgfqpoint{-0.000000in}{0.000000in}}{%
\pgfpathmoveto{\pgfqpoint{-0.000000in}{0.000000in}}%
\pgfpathlineto{\pgfqpoint{-0.048611in}{0.000000in}}%
\pgfusepath{stroke,fill}%
}%
\begin{pgfscope}%
\pgfsys@transformshift{0.708026in}{0.602918in}%
\pgfsys@useobject{currentmarker}{}%
\end{pgfscope}%
\end{pgfscope}%
\begin{pgfscope}%
\definecolor{textcolor}{rgb}{0.000000,0.000000,0.000000}%
\pgfsetstrokecolor{textcolor}%
\pgfsetfillcolor{textcolor}%
\pgftext[x=0.294444in, y=0.554693in, left, base]{\color{textcolor}{\sffamily\fontsize{10.000000}{12.000000}\selectfont\catcode`\^=\active\def^{\ifmmode\sp\else\^{}\fi}\catcode`\%=\active\def
\end{pgfscope}%
\begin{pgfscope}%
\pgfsetbuttcap%
\pgfsetroundjoin%
\definecolor{currentfill}{rgb}{0.000000,0.000000,0.000000}%
\pgfsetfillcolor{currentfill}%
\pgfsetlinewidth{0.803000pt}%
\definecolor{currentstroke}{rgb}{0.000000,0.000000,0.000000}%
\pgfsetstrokecolor{currentstroke}%
\pgfsetdash{}{0pt}%
\pgfsys@defobject{currentmarker}{\pgfqpoint{-0.048611in}{0.000000in}}{\pgfqpoint{-0.000000in}{0.000000in}}{%
\pgfpathmoveto{\pgfqpoint{-0.000000in}{0.000000in}}%
\pgfpathlineto{\pgfqpoint{-0.048611in}{0.000000in}}%
\pgfusepath{stroke,fill}%
}%
\begin{pgfscope}%
\pgfsys@transformshift{0.708026in}{0.981344in}%
\pgfsys@useobject{currentmarker}{}%
\end{pgfscope}%
\end{pgfscope}%
\begin{pgfscope}%
\definecolor{textcolor}{rgb}{0.000000,0.000000,0.000000}%
\pgfsetstrokecolor{textcolor}%
\pgfsetfillcolor{textcolor}%
\pgftext[x=0.294444in, y=0.933118in, left, base]{\color{textcolor}{\sffamily\fontsize{10.000000}{12.000000}\selectfont\catcode`\^=\active\def^{\ifmmode\sp\else\^{}\fi}\catcode`\%=\active\def
\end{pgfscope}%
\begin{pgfscope}%
\pgfsetbuttcap%
\pgfsetroundjoin%
\definecolor{currentfill}{rgb}{0.000000,0.000000,0.000000}%
\pgfsetfillcolor{currentfill}%
\pgfsetlinewidth{0.803000pt}%
\definecolor{currentstroke}{rgb}{0.000000,0.000000,0.000000}%
\pgfsetstrokecolor{currentstroke}%
\pgfsetdash{}{0pt}%
\pgfsys@defobject{currentmarker}{\pgfqpoint{-0.048611in}{0.000000in}}{\pgfqpoint{-0.000000in}{0.000000in}}{%
\pgfpathmoveto{\pgfqpoint{-0.000000in}{0.000000in}}%
\pgfpathlineto{\pgfqpoint{-0.048611in}{0.000000in}}%
\pgfusepath{stroke,fill}%
}%
\begin{pgfscope}%
\pgfsys@transformshift{0.708026in}{1.359769in}%
\pgfsys@useobject{currentmarker}{}%
\end{pgfscope}%
\end{pgfscope}%
\begin{pgfscope}%
\definecolor{textcolor}{rgb}{0.000000,0.000000,0.000000}%
\pgfsetstrokecolor{textcolor}%
\pgfsetfillcolor{textcolor}%
\pgftext[x=0.294444in, y=1.311544in, left, base]{\color{textcolor}{\sffamily\fontsize{10.000000}{12.000000}\selectfont\catcode`\^=\active\def^{\ifmmode\sp\else\^{}\fi}\catcode`\%=\active\def
\end{pgfscope}%
\begin{pgfscope}%
\pgfsetbuttcap%
\pgfsetroundjoin%
\definecolor{currentfill}{rgb}{0.000000,0.000000,0.000000}%
\pgfsetfillcolor{currentfill}%
\pgfsetlinewidth{0.803000pt}%
\definecolor{currentstroke}{rgb}{0.000000,0.000000,0.000000}%
\pgfsetstrokecolor{currentstroke}%
\pgfsetdash{}{0pt}%
\pgfsys@defobject{currentmarker}{\pgfqpoint{-0.048611in}{0.000000in}}{\pgfqpoint{-0.000000in}{0.000000in}}{%
\pgfpathmoveto{\pgfqpoint{-0.000000in}{0.000000in}}%
\pgfpathlineto{\pgfqpoint{-0.048611in}{0.000000in}}%
\pgfusepath{stroke,fill}%
}%
\begin{pgfscope}%
\pgfsys@transformshift{0.708026in}{1.738194in}%
\pgfsys@useobject{currentmarker}{}%
\end{pgfscope}%
\end{pgfscope}%
\begin{pgfscope}%
\definecolor{textcolor}{rgb}{0.000000,0.000000,0.000000}%
\pgfsetstrokecolor{textcolor}%
\pgfsetfillcolor{textcolor}%
\pgftext[x=0.541359in, y=1.689969in, left, base]{\color{textcolor}{\sffamily\fontsize{10.000000}{12.000000}\selectfont\catcode`\^=\active\def^{\ifmmode\sp\else\^{}\fi}\catcode`\%=\active\def
\end{pgfscope}%
\begin{pgfscope}%
\pgfsetbuttcap%
\pgfsetroundjoin%
\definecolor{currentfill}{rgb}{0.000000,0.000000,0.000000}%
\pgfsetfillcolor{currentfill}%
\pgfsetlinewidth{0.803000pt}%
\definecolor{currentstroke}{rgb}{0.000000,0.000000,0.000000}%
\pgfsetstrokecolor{currentstroke}%
\pgfsetdash{}{0pt}%
\pgfsys@defobject{currentmarker}{\pgfqpoint{-0.048611in}{0.000000in}}{\pgfqpoint{-0.000000in}{0.000000in}}{%
\pgfpathmoveto{\pgfqpoint{-0.000000in}{0.000000in}}%
\pgfpathlineto{\pgfqpoint{-0.048611in}{0.000000in}}%
\pgfusepath{stroke,fill}%
}%
\begin{pgfscope}%
\pgfsys@transformshift{0.708026in}{2.116620in}%
\pgfsys@useobject{currentmarker}{}%
\end{pgfscope}%
\end{pgfscope}%
\begin{pgfscope}%
\definecolor{textcolor}{rgb}{0.000000,0.000000,0.000000}%
\pgfsetstrokecolor{textcolor}%
\pgfsetfillcolor{textcolor}%
\pgftext[x=0.402469in, y=2.068395in, left, base]{\color{textcolor}{\sffamily\fontsize{10.000000}{12.000000}\selectfont\catcode`\^=\active\def^{\ifmmode\sp\else\^{}\fi}\catcode`\%=\active\def
\end{pgfscope}%
\begin{pgfscope}%
\pgfsetbuttcap%
\pgfsetroundjoin%
\definecolor{currentfill}{rgb}{0.000000,0.000000,0.000000}%
\pgfsetfillcolor{currentfill}%
\pgfsetlinewidth{0.803000pt}%
\definecolor{currentstroke}{rgb}{0.000000,0.000000,0.000000}%
\pgfsetstrokecolor{currentstroke}%
\pgfsetdash{}{0pt}%
\pgfsys@defobject{currentmarker}{\pgfqpoint{-0.048611in}{0.000000in}}{\pgfqpoint{-0.000000in}{0.000000in}}{%
\pgfpathmoveto{\pgfqpoint{-0.000000in}{0.000000in}}%
\pgfpathlineto{\pgfqpoint{-0.048611in}{0.000000in}}%
\pgfusepath{stroke,fill}%
}%
\begin{pgfscope}%
\pgfsys@transformshift{0.708026in}{2.495045in}%
\pgfsys@useobject{currentmarker}{}%
\end{pgfscope}%
\end{pgfscope}%
\begin{pgfscope}%
\definecolor{textcolor}{rgb}{0.000000,0.000000,0.000000}%
\pgfsetstrokecolor{textcolor}%
\pgfsetfillcolor{textcolor}%
\pgftext[x=0.402469in, y=2.446820in, left, base]{\color{textcolor}{\sffamily\fontsize{10.000000}{12.000000}\selectfont\catcode`\^=\active\def^{\ifmmode\sp\else\^{}\fi}\catcode`\%=\active\def
\end{pgfscope}%
\begin{pgfscope}%
\pgfsetbuttcap%
\pgfsetroundjoin%
\definecolor{currentfill}{rgb}{0.000000,0.000000,0.000000}%
\pgfsetfillcolor{currentfill}%
\pgfsetlinewidth{0.803000pt}%
\definecolor{currentstroke}{rgb}{0.000000,0.000000,0.000000}%
\pgfsetstrokecolor{currentstroke}%
\pgfsetdash{}{0pt}%
\pgfsys@defobject{currentmarker}{\pgfqpoint{-0.048611in}{0.000000in}}{\pgfqpoint{-0.000000in}{0.000000in}}{%
\pgfpathmoveto{\pgfqpoint{-0.000000in}{0.000000in}}%
\pgfpathlineto{\pgfqpoint{-0.048611in}{0.000000in}}%
\pgfusepath{stroke,fill}%
}%
\begin{pgfscope}%
\pgfsys@transformshift{0.708026in}{2.873471in}%
\pgfsys@useobject{currentmarker}{}%
\end{pgfscope}%
\end{pgfscope}%
\begin{pgfscope}%
\definecolor{textcolor}{rgb}{0.000000,0.000000,0.000000}%
\pgfsetstrokecolor{textcolor}%
\pgfsetfillcolor{textcolor}%
\pgftext[x=0.402469in, y=2.825246in, left, base]{\color{textcolor}{\sffamily\fontsize{10.000000}{12.000000}\selectfont\catcode`\^=\active\def^{\ifmmode\sp\else\^{}\fi}\catcode`\%=\active\def
\end{pgfscope}%
\begin{pgfscope}%
\definecolor{textcolor}{rgb}{0.000000,0.000000,0.000000}%
\pgfsetstrokecolor{textcolor}%
\pgfsetfillcolor{textcolor}%
\pgftext[x=0.238889in,y=1.735748in,,bottom,rotate=90.000000]{\color{textcolor}{\sffamily\fontsize{10.000000}{12.000000}\selectfont\catcode`\^=\active\def^{\ifmmode\sp\else\^{}\fi}\catcode`\%=\active\def
\end{pgfscope}%
\begin{pgfscope}%
\pgfsetrectcap%
\pgfsetmiterjoin%
\pgfsetlinewidth{0.803000pt}%
\definecolor{currentstroke}{rgb}{0.000000,0.000000,0.000000}%
\pgfsetstrokecolor{currentstroke}%
\pgfsetdash{}{0pt}%
\pgfpathmoveto{\pgfqpoint{0.708026in}{0.515123in}}%
\pgfpathlineto{\pgfqpoint{0.708026in}{2.956373in}}%
\pgfusepath{stroke}%
\end{pgfscope}%
\begin{pgfscope}%
\pgfsetrectcap%
\pgfsetmiterjoin%
\pgfsetlinewidth{0.803000pt}%
\definecolor{currentstroke}{rgb}{0.000000,0.000000,0.000000}%
\pgfsetstrokecolor{currentstroke}%
\pgfsetdash{}{0pt}%
\pgfpathmoveto{\pgfqpoint{3.149276in}{0.515123in}}%
\pgfpathlineto{\pgfqpoint{3.149276in}{2.956373in}}%
\pgfusepath{stroke}%
\end{pgfscope}%
\begin{pgfscope}%
\pgfsetrectcap%
\pgfsetmiterjoin%
\pgfsetlinewidth{0.803000pt}%
\definecolor{currentstroke}{rgb}{0.000000,0.000000,0.000000}%
\pgfsetstrokecolor{currentstroke}%
\pgfsetdash{}{0pt}%
\pgfpathmoveto{\pgfqpoint{0.708026in}{0.515123in}}%
\pgfpathlineto{\pgfqpoint{3.149276in}{0.515123in}}%
\pgfusepath{stroke}%
\end{pgfscope}%
\begin{pgfscope}%
\pgfsetrectcap%
\pgfsetmiterjoin%
\pgfsetlinewidth{0.803000pt}%
\definecolor{currentstroke}{rgb}{0.000000,0.000000,0.000000}%
\pgfsetstrokecolor{currentstroke}%
\pgfsetdash{}{0pt}%
\pgfpathmoveto{\pgfqpoint{0.708026in}{2.956373in}}%
\pgfpathlineto{\pgfqpoint{3.149276in}{2.956373in}}%
\pgfusepath{stroke}%
\end{pgfscope}%
\begin{pgfscope}%
\pgfsetbuttcap%
\pgfsetmiterjoin%
\definecolor{currentfill}{rgb}{1.000000,1.000000,1.000000}%
\pgfsetfillcolor{currentfill}%
\pgfsetfillopacity{0.800000}%
\pgfsetlinewidth{1.003750pt}%
\definecolor{currentstroke}{rgb}{0.800000,0.800000,0.800000}%
\pgfsetstrokecolor{currentstroke}%
\pgfsetstrokeopacity{0.800000}%
\pgfsetdash{}{0pt}%
\pgfpathmoveto{\pgfqpoint{2.291438in}{1.937854in}}%
\pgfpathlineto{\pgfqpoint{3.071498in}{1.937854in}}%
\pgfpathquadraticcurveto{\pgfqpoint{3.093720in}{1.937854in}}{\pgfqpoint{3.093720in}{1.960077in}}%
\pgfpathlineto{\pgfqpoint{3.093720in}{2.878596in}}%
\pgfpathquadraticcurveto{\pgfqpoint{3.093720in}{2.900818in}}{\pgfqpoint{3.071498in}{2.900818in}}%
\pgfpathlineto{\pgfqpoint{2.291438in}{2.900818in}}%
\pgfpathquadraticcurveto{\pgfqpoint{2.269216in}{2.900818in}}{\pgfqpoint{2.269216in}{2.878596in}}%
\pgfpathlineto{\pgfqpoint{2.269216in}{1.960077in}}%
\pgfpathquadraticcurveto{\pgfqpoint{2.269216in}{1.937854in}}{\pgfqpoint{2.291438in}{1.937854in}}%
\pgfpathlineto{\pgfqpoint{2.291438in}{1.937854in}}%
\pgfpathclose%
\pgfusepath{stroke,fill}%
\end{pgfscope}%
\begin{pgfscope}%
\pgfsetbuttcap%
\pgfsetmiterjoin%
\definecolor{currentfill}{rgb}{0.000000,0.000000,1.000000}%
\pgfsetfillcolor{currentfill}%
\pgfsetlinewidth{1.003750pt}%
\definecolor{currentstroke}{rgb}{0.000000,0.000000,1.000000}%
\pgfsetstrokecolor{currentstroke}%
\pgfsetdash{}{0pt}%
\pgfpathmoveto{\pgfqpoint{2.313660in}{2.778596in}}%
\pgfpathlineto{\pgfqpoint{2.535883in}{2.778596in}}%
\pgfpathlineto{\pgfqpoint{2.535883in}{2.856373in}}%
\pgfpathlineto{\pgfqpoint{2.313660in}{2.856373in}}%
\pgfpathlineto{\pgfqpoint{2.313660in}{2.778596in}}%
\pgfpathclose%
\pgfusepath{stroke,fill}%
\end{pgfscope}%
\begin{pgfscope}%
\definecolor{textcolor}{rgb}{0.000000,0.000000,0.000000}%
\pgfsetstrokecolor{textcolor}%
\pgfsetfillcolor{textcolor}%
\pgftext[x=2.624772in,y=2.778596in,left,base]{\color{textcolor}{\sffamily\fontsize{8.000000}{9.600000}\selectfont\catcode`\^=\active\def^{\ifmmode\sp\else\^{}\fi}\catcode`\%=\active\def
\end{pgfscope}%
\begin{pgfscope}%
\pgfsetbuttcap%
\pgfsetmiterjoin%
\definecolor{currentfill}{rgb}{0.000000,0.501961,0.000000}%
\pgfsetfillcolor{currentfill}%
\pgfsetlinewidth{1.003750pt}%
\definecolor{currentstroke}{rgb}{0.000000,0.501961,0.000000}%
\pgfsetstrokecolor{currentstroke}%
\pgfsetdash{}{0pt}%
\pgfpathmoveto{\pgfqpoint{2.313660in}{2.623657in}}%
\pgfpathlineto{\pgfqpoint{2.535883in}{2.623657in}}%
\pgfpathlineto{\pgfqpoint{2.535883in}{2.701435in}}%
\pgfpathlineto{\pgfqpoint{2.313660in}{2.701435in}}%
\pgfpathlineto{\pgfqpoint{2.313660in}{2.623657in}}%
\pgfpathclose%
\pgfusepath{stroke,fill}%
\end{pgfscope}%
\begin{pgfscope}%
\definecolor{textcolor}{rgb}{0.000000,0.000000,0.000000}%
\pgfsetstrokecolor{textcolor}%
\pgfsetfillcolor{textcolor}%
\pgftext[x=2.624772in,y=2.623657in,left,base]{\color{textcolor}{\sffamily\fontsize{8.000000}{9.600000}\selectfont\catcode`\^=\active\def^{\ifmmode\sp\else\^{}\fi}\catcode`\%=\active\def
\end{pgfscope}%
\begin{pgfscope}%
\pgfsetbuttcap%
\pgfsetmiterjoin%
\definecolor{currentfill}{rgb}{1.000000,0.000000,1.000000}%
\pgfsetfillcolor{currentfill}%
\pgfsetlinewidth{1.003750pt}%
\definecolor{currentstroke}{rgb}{1.000000,0.000000,1.000000}%
\pgfsetstrokecolor{currentstroke}%
\pgfsetdash{}{0pt}%
\pgfpathmoveto{\pgfqpoint{2.313660in}{2.468719in}}%
\pgfpathlineto{\pgfqpoint{2.535883in}{2.468719in}}%
\pgfpathlineto{\pgfqpoint{2.535883in}{2.546497in}}%
\pgfpathlineto{\pgfqpoint{2.313660in}{2.546497in}}%
\pgfpathlineto{\pgfqpoint{2.313660in}{2.468719in}}%
\pgfpathclose%
\pgfusepath{stroke,fill}%
\end{pgfscope}%
\begin{pgfscope}%
\definecolor{textcolor}{rgb}{0.000000,0.000000,0.000000}%
\pgfsetstrokecolor{textcolor}%
\pgfsetfillcolor{textcolor}%
\pgftext[x=2.624772in,y=2.468719in,left,base]{\color{textcolor}{\sffamily\fontsize{8.000000}{9.600000}\selectfont\catcode`\^=\active\def^{\ifmmode\sp\else\^{}\fi}\catcode`\%=\active\def
\end{pgfscope}%
\begin{pgfscope}%
\pgfsetbuttcap%
\pgfsetmiterjoin%
\definecolor{currentfill}{rgb}{0.000000,0.000000,0.000000}%
\pgfsetfillcolor{currentfill}%
\pgfsetlinewidth{1.003750pt}%
\definecolor{currentstroke}{rgb}{0.000000,0.000000,0.000000}%
\pgfsetstrokecolor{currentstroke}%
\pgfsetdash{}{0pt}%
\pgfpathmoveto{\pgfqpoint{2.313660in}{2.313781in}}%
\pgfpathlineto{\pgfqpoint{2.535883in}{2.313781in}}%
\pgfpathlineto{\pgfqpoint{2.535883in}{2.391558in}}%
\pgfpathlineto{\pgfqpoint{2.313660in}{2.391558in}}%
\pgfpathlineto{\pgfqpoint{2.313660in}{2.313781in}}%
\pgfpathclose%
\pgfusepath{stroke,fill}%
\end{pgfscope}%
\begin{pgfscope}%
\definecolor{textcolor}{rgb}{0.000000,0.000000,0.000000}%
\pgfsetstrokecolor{textcolor}%
\pgfsetfillcolor{textcolor}%
\pgftext[x=2.624772in,y=2.313781in,left,base]{\color{textcolor}{\sffamily\fontsize{8.000000}{9.600000}\selectfont\catcode`\^=\active\def^{\ifmmode\sp\else\^{}\fi}\catcode`\%=\active\def
\end{pgfscope}%
\begin{pgfscope}%
\pgfsetbuttcap%
\pgfsetmiterjoin%
\definecolor{currentfill}{rgb}{1.000000,0.000000,0.000000}%
\pgfsetfillcolor{currentfill}%
\pgfsetlinewidth{1.003750pt}%
\definecolor{currentstroke}{rgb}{1.000000,0.000000,0.000000}%
\pgfsetstrokecolor{currentstroke}%
\pgfsetdash{}{0pt}%
\pgfpathmoveto{\pgfqpoint{2.313660in}{2.158842in}}%
\pgfpathlineto{\pgfqpoint{2.535883in}{2.158842in}}%
\pgfpathlineto{\pgfqpoint{2.535883in}{2.236620in}}%
\pgfpathlineto{\pgfqpoint{2.313660in}{2.236620in}}%
\pgfpathlineto{\pgfqpoint{2.313660in}{2.158842in}}%
\pgfpathclose%
\pgfusepath{stroke,fill}%
\end{pgfscope}%
\begin{pgfscope}%
\definecolor{textcolor}{rgb}{0.000000,0.000000,0.000000}%
\pgfsetstrokecolor{textcolor}%
\pgfsetfillcolor{textcolor}%
\pgftext[x=2.624772in,y=2.158842in,left,base]{\color{textcolor}{\sffamily\fontsize{8.000000}{9.600000}\selectfont\catcode`\^=\active\def^{\ifmmode\sp\else\^{}\fi}\catcode`\%=\active\def
\end{pgfscope}%
\begin{pgfscope}%
\pgfsetrectcap%
\pgfsetroundjoin%
\pgfsetlinewidth{1.505625pt}%
\definecolor{currentstroke}{rgb}{1.000000,0.647059,0.000000}%
\pgfsetstrokecolor{currentstroke}%
\pgfsetdash{}{0pt}%
\pgfpathmoveto{\pgfqpoint{2.313660in}{2.042793in}}%
\pgfpathlineto{\pgfqpoint{2.424772in}{2.042793in}}%
\pgfpathlineto{\pgfqpoint{2.535883in}{2.042793in}}%
\pgfusepath{stroke}%
\end{pgfscope}%
\begin{pgfscope}%
\pgfsetbuttcap%
\pgfsetmiterjoin%
\definecolor{currentfill}{rgb}{1.000000,0.647059,0.000000}%
\pgfsetfillcolor{currentfill}%
\pgfsetlinewidth{1.003750pt}%
\definecolor{currentstroke}{rgb}{1.000000,0.647059,0.000000}%
\pgfsetstrokecolor{currentstroke}%
\pgfsetdash{}{0pt}%
\pgfsys@defobject{currentmarker}{\pgfqpoint{-0.055556in}{-0.055556in}}{\pgfqpoint{0.055556in}{0.055556in}}{%
\pgfpathmoveto{\pgfqpoint{0.000000in}{0.055556in}}%
\pgfpathlineto{\pgfqpoint{-0.055556in}{-0.055556in}}%
\pgfpathlineto{\pgfqpoint{0.055556in}{-0.055556in}}%
\pgfpathlineto{\pgfqpoint{0.000000in}{0.055556in}}%
\pgfpathclose%
\pgfusepath{stroke,fill}%
}%
\begin{pgfscope}%
\pgfsys@transformshift{2.424772in}{2.042793in}%
\pgfsys@useobject{currentmarker}{}%
\end{pgfscope}%
\end{pgfscope}%
\begin{pgfscope}%
\definecolor{textcolor}{rgb}{0.000000,0.000000,0.000000}%
\pgfsetstrokecolor{textcolor}%
\pgfsetfillcolor{textcolor}%
\pgftext[x=2.624772in,y=2.003904in,left,base]{\color{textcolor}{\sffamily\fontsize{8.000000}{9.600000}\selectfont\catcode`\^=\active\def^{\ifmmode\sp\else\^{}\fi}\catcode`\%=\active\def
\end{pgfscope}%
\end{pgfpicture}%
\makeatother%
\endgroup%

%% file: figures/tex/prediction-heatmaps/prediction-heatmaps_140_time1657876089.234451.tex
\begingroup%
\makeatletter%
\begin{pgfpicture}%
\pgfpathrectangle{\pgfpointorigin}{\pgfqpoint{2.756539in}{1.627760in}}%
\pgfusepath{use as bounding box, clip}%
\begin{pgfscope}%
\pgfsetbuttcap%
\pgfsetmiterjoin%
\definecolor{currentfill}{rgb}{1.000000,1.000000,1.000000}%
\pgfsetfillcolor{currentfill}%
\pgfsetlinewidth{0.000000pt}%
\definecolor{currentstroke}{rgb}{1.000000,1.000000,1.000000}%
\pgfsetstrokecolor{currentstroke}%
\pgfsetdash{}{0pt}%
\pgfpathmoveto{\pgfqpoint{0.000000in}{0.000000in}}%
\pgfpathlineto{\pgfqpoint{2.756539in}{0.000000in}}%
\pgfpathlineto{\pgfqpoint{2.756539in}{1.627760in}}%
\pgfpathlineto{\pgfqpoint{0.000000in}{1.627760in}}%
\pgfpathlineto{\pgfqpoint{0.000000in}{0.000000in}}%
\pgfpathclose%
\pgfusepath{fill}%
\end{pgfscope}%
\begin{pgfscope}%
\pgfsetbuttcap%
\pgfsetmiterjoin%
\definecolor{currentfill}{rgb}{1.000000,1.000000,1.000000}%
\pgfsetfillcolor{currentfill}%
\pgfsetlinewidth{0.000000pt}%
\definecolor{currentstroke}{rgb}{0.000000,0.000000,0.000000}%
\pgfsetstrokecolor{currentstroke}%
\pgfsetstrokeopacity{0.000000}%
\pgfsetdash{}{0pt}%
\pgfpathmoveto{\pgfqpoint{0.215289in}{0.100000in}}%
\pgfpathlineto{\pgfqpoint{2.656539in}{0.100000in}}%
\pgfpathlineto{\pgfqpoint{2.656539in}{1.079395in}}%
\pgfpathlineto{\pgfqpoint{0.215289in}{1.079395in}}%
\pgfpathlineto{\pgfqpoint{0.215289in}{0.100000in}}%
\pgfpathclose%
\pgfusepath{fill}%
\end{pgfscope}%
\begin{pgfscope}%
\pgfpathrectangle{\pgfqpoint{0.215289in}{0.100000in}}{\pgfqpoint{2.441250in}{0.979395in}}%
\pgfusepath{clip}%
\pgfsys@transformshift{0.215289in}{0.100000in}%
\pgftext[left,bottom]{\includegraphics[interpolate=true,width=2.450000in,height=0.980000in]{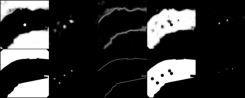}}%
\end{pgfscope}%
\begin{pgfscope}%
\pgfsetrectcap%
\pgfsetmiterjoin%
\pgfsetlinewidth{0.803000pt}%
\definecolor{currentstroke}{rgb}{0.000000,0.000000,0.000000}%
\pgfsetstrokecolor{currentstroke}%
\pgfsetdash{}{0pt}%
\pgfpathmoveto{\pgfqpoint{0.215289in}{0.100000in}}%
\pgfpathlineto{\pgfqpoint{0.215289in}{1.079395in}}%
\pgfusepath{stroke}%
\end{pgfscope}%
\begin{pgfscope}%
\pgfsetrectcap%
\pgfsetmiterjoin%
\pgfsetlinewidth{0.803000pt}%
\definecolor{currentstroke}{rgb}{0.000000,0.000000,0.000000}%
\pgfsetstrokecolor{currentstroke}%
\pgfsetdash{}{0pt}%
\pgfpathmoveto{\pgfqpoint{2.656539in}{0.100000in}}%
\pgfpathlineto{\pgfqpoint{2.656539in}{1.079395in}}%
\pgfusepath{stroke}%
\end{pgfscope}%
\begin{pgfscope}%
\pgfsetrectcap%
\pgfsetmiterjoin%
\pgfsetlinewidth{0.803000pt}%
\definecolor{currentstroke}{rgb}{0.000000,0.000000,0.000000}%
\pgfsetstrokecolor{currentstroke}%
\pgfsetdash{}{0pt}%
\pgfpathmoveto{\pgfqpoint{0.215289in}{0.100000in}}%
\pgfpathlineto{\pgfqpoint{2.656539in}{0.100000in}}%
\pgfusepath{stroke}%
\end{pgfscope}%
\begin{pgfscope}%
\pgfsetrectcap%
\pgfsetmiterjoin%
\pgfsetlinewidth{0.803000pt}%
\definecolor{currentstroke}{rgb}{0.000000,0.000000,0.000000}%
\pgfsetstrokecolor{currentstroke}%
\pgfsetdash{}{0pt}%
\pgfpathmoveto{\pgfqpoint{0.215289in}{1.079395in}}%
\pgfpathlineto{\pgfqpoint{2.656539in}{1.079395in}}%
\pgfusepath{stroke}%
\end{pgfscope}%
\begin{pgfscope}%
\definecolor{textcolor}{rgb}{0.000000,0.000000,0.000000}%
\pgfsetstrokecolor{textcolor}%
\pgfsetfillcolor{textcolor}%
\pgftext[x=0.397554in, y=1.139214in, left, base,rotate=45.000000]{\color{textcolor}{\rmfamily\fontsize{8.000000}{9.600000}\selectfont\catcode`\^=\active\def^{\ifmmode\sp\else\^{}\fi}\catcode`\%=\active\def
\end{pgfscope}%
\begin{pgfscope}%
\definecolor{textcolor}{rgb}{0.000000,0.000000,0.000000}%
\pgfsetstrokecolor{textcolor}%
\pgfsetfillcolor{textcolor}%
\pgftext[x=0.872119in, y=1.139214in, left, base,rotate=45.000000]{\color{textcolor}{\rmfamily\fontsize{8.000000}{9.600000}\selectfont\catcode`\^=\active\def^{\ifmmode\sp\else\^{}\fi}\catcode`\%=\active\def
\end{pgfscope}%
\begin{pgfscope}%
\definecolor{textcolor}{rgb}{0.000000,0.000000,0.000000}%
\pgfsetstrokecolor{textcolor}%
\pgfsetfillcolor{textcolor}%
\pgftext[x=1.308251in, y=1.139214in, left, base,rotate=45.000000]{\color{textcolor}{\rmfamily\fontsize{8.000000}{9.600000}\selectfont\catcode`\^=\active\def^{\ifmmode\sp\else\^{}\fi}\catcode`\%=\active\def
\end{pgfscope}%
\begin{pgfscope}%
\definecolor{textcolor}{rgb}{0.000000,0.000000,0.000000}%
\pgfsetstrokecolor{textcolor}%
\pgfsetfillcolor{textcolor}%
\pgftext[x=1.856132in, y=1.139214in, left, base,rotate=45.000000]{\color{textcolor}{\rmfamily\fontsize{8.000000}{9.600000}\selectfont\catcode`\^=\active\def^{\ifmmode\sp\else\^{}\fi}\catcode`\%=\active\def
\end{pgfscope}%
\begin{pgfscope}%
\definecolor{textcolor}{rgb}{0.000000,0.000000,0.000000}%
\pgfsetstrokecolor{textcolor}%
\pgfsetfillcolor{textcolor}%
\pgftext[x=2.323699in, y=1.139214in, left, base,rotate=45.000000]{\color{textcolor}{\rmfamily\fontsize{8.000000}{9.600000}\selectfont\catcode`\^=\active\def^{\ifmmode\sp\else\^{}\fi}\catcode`\%=\active\def
\end{pgfscope}%
\begin{pgfscope}%
\definecolor{textcolor}{rgb}{0.000000,0.000000,0.000000}%
\pgfsetstrokecolor{textcolor}%
\pgfsetfillcolor{textcolor}%
\pgftext[x=0.136539in,y=0.748300in,,]{\color{textcolor}{\rmfamily\fontsize{10.000000}{12.000000}\selectfont\catcode`\^=\active\def^{\ifmmode\sp\else\^{}\fi}\catcode`\%=\active\def
\end{pgfscope}%
\begin{pgfscope}%
\definecolor{textcolor}{rgb}{0.000000,0.000000,0.000000}%
\pgfsetstrokecolor{textcolor}%
\pgfsetfillcolor{textcolor}%
\pgftext[x=0.136539in,y=0.330925in,,]{\color{textcolor}{\rmfamily\fontsize{10.000000}{12.000000}\selectfont\catcode`\^=\active\def^{\ifmmode\sp\else\^{}\fi}\catcode`\%=\active\def
\end{pgfscope}%
\end{pgfpicture}%
\makeatother%
\endgroup%

%% file: chapters/proposed_approach/solution_architecture.tex
\subsection{Solution architecture}\label{subseq:sol_arch}
The key components of our solution architecture are \begin{enumerate*}[label=\alph*)]
\item a pseudo-camera view pre-processing method $G$ that rasterizes sparse and unstructured point clouds to structured and dense virtual camera views,
\item the modality specific feature encoders,
\item the positional-ray encodings,
\item a transform and fuse module that generates BEV features from both actual and virtual camera views,
\item a decoder that upscales the BEV features to the final BEV map shape. 
\end{enumerate*}

\textbf{LiDAR preprocessing:}

\begin{figure}[ht]
    \centering
    \import{\paperpath/figures/tex/lidar-channels-pca}{lidar-channels-pca_20231219_sample_140_time1657878183.065121.tex}
    \caption{PCA (n=1) of a pseudo-camera lidar point cloud $\pseudopcd$ (see also \cref{fig:lidar-features-pillars}).}
    \label{fig:lidar-features}
\end{figure}

\begin{figure}[ptb]
	\centering
	\includegraphics[width=0.7\columnwidth]{\paperpath/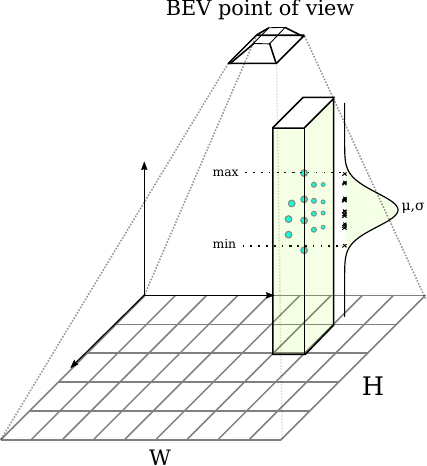}
	\caption{Pre-processing $\mathcal{G}$ of LiDAR point cloud data. The points are projected on a BEV grid defined by the view a BEV point of view and the \(z=0\) ground place. Each cell contains statistics of the z-coordinate values of the points that fall within the cell (see also \cref{fig:lidar-features}).}
	\label{fig:lidar-features-pillars}
\end{figure}

Raw point-clouds are unstructured and sparse, and hence not compatible with most deep learning processing blocks, that expect structured vectors of defined shapes. Inspired by \cite{yang2018pixor,lang2019pointpillars,beltran2018birdnet}, we pre-process LiDAR point clouds and conver them to pseudo-images, enabling their processing with convolution architectures.

Let $\rawpcd$ represent a 3D point cloud with $N_{\rawpcd}$ points

\[\rawpcd=\{(x_n,y_n,z_n) : i\in \naturals, n\leq N_{\rawpcd} \}\] 

Let also $\pseudopcd$ represent the processed pseudo-image of $\rawpcd$
\[\pseudopcd = \mathcal{G}(\rawpcd) \in \reals^{H \times W \times C_S}\]  

where
\begin{description}[left=15pt]
    \item[$\mathcal{G}$:] $\reals^3 \rightarrow \reals^{N_\text{LiDAR} \times C_\text{LiDAR}}$
    
    \item[$\pseudopcd$:] $\{ f_{ij} : i \in \setindex{H} , \, j \in \setindex{W} \}$
    
    \item[$f_{ij}$:] $\begin{pmatrix}
    \expectation{\voxel_{ij}} & \variance{\voxel_{ij}} & \max(\voxel_{ij}) & \min(\voxel_{ij})
    \end{pmatrix}^T$
    
    \item[$\voxel_{ij}$:] Aggregated points within each voxel given by \cref{eq:agg-voxels}
    
    \item[$\floor{\cdot}$:] Rounds down to the closest integer
    
    \item[$H, W$:] Pseudo-image dimensions, such that $HW = N_\text{LiDAR}$ in \cref{eq:heterogeneous-data} notation
    
    \item[$d_\pseudopcd$:] Spatial dimension of the pseudo-image grid (see \cref{fig:lidar-features-pillars})
\end{description}

\begin{equation}
\voxel_{ij} = \{ z_n  | \, (x_n,y_n,z_n) \in  \rawpcd | \floor{\frac{x_n}{d_\pseudopcd}}=i \land \floor{\frac{y_n}{d_\pseudopcd}}=j \}
    \label{eq:agg-voxels}
\end{equation}

In our application, the sparse modality corresponds to the LiDAR sensors, therefore we introduce the $\{\cdot\}_\text{LiDAR}$ subscript to ground our notation. We preprocess and standardize the unstructured and irregularly sized point clouds $\rawpcd_{\text{LiDAR}, u,\tau}$ to their structured and uniform BEV pseudo-image representations $\pseudopcd_{\text{LiDAR}, u,\tau}$, while preserving 3D geometry features through points statistics. This processing is illustrated in \cref{fig:lidar-features-pillars}. Finally, we aggregate processed pseudo-images for all $V_\text{LiDAR}$ lidar sensors and time instances $t$, to form $\pseudopcd_\text{LiDAR} \in \reals^{t \times V_\text{LiDAR} \times N_\text{LiDAR} \times C_\text{LiDAR}}$, matching \cref{eq:heterogeneous-data} notation.

\[\pseudopcd_\text{LiDAR} = \{\mathcal{G}(\rawpcd_{\text{LiDAR},v,\tau}) : v,\tau \in \mathbb{N} \mid v \leq V_\text{LiDAR} , \, \tau \leq t \} \]

\textbf{Modality encoders:}

We revisit \cref{eq:heterogeneous-data}, by introducing further grounding subscript notation to the dense modalities.
\[\Vset =\{\Dele_{\text{RGB}},\Dele_{\text{LWIR}} \} \cup \{\Sele_{\text{LiDAR}}\} \]

\begin{itemize}
    \item \(\Dele_{\text{RGB}},\Dele_{\text{LWIR}} \in \reals^{t \times \Vset_D \times H\times W \times 3}\) represent dense monocular camera views,
    \item \(\pseudopcd_{\text{LiDAR}}  \in  \reals^{t \times V_\text{LiDAR} \times N_\text{LiDAR} \times C_\text{LiDAR}} \) represents the raw and sparse 3D point-cloud of lidar $v$ at time instant $\tau$.
\end{itemize}

We encode modalities into latent features, using modality-specific backbones at different resolutions. In order to extract features from RGB and LWIR camera views, we convert LWIR images to RGB, and extract feature maps from the intermediate layers of a shared pre-trained EfficientNet-b4. Since we can't use transfer learning to extract features from the LiDAR pseudo-views \(\rawpcd_i\), we use a randomly initialized U-Net \cite{ronneberger2015unet} feature extractor, similarly to \cite{hendy2020fishing}. Modality encoders are projecting their features to a common dimension, hence we adopt the notation feature $f^{\tau,k}_{i} \in \reals^{d_e}$ as the feature extracted from camera or pseudo-camera $k$ at time index $\tau$.

\textbf{Cross attention}
For any world point \(x^W \in \reals^3 \), there exists a corresponding image pixel coordinate \(x^I = \begin{psmallmatrix}
p_i,p_j,d
\end{psmallmatrix}\).
For a view \(k\), world points are connected to camera coordinates by \cref{eq:camera-projection}:
\begin{equation}
x^I = I_k E_k (x^W - \camposition_k)
    \label{eq:camera-projection}
\end{equation}

Where \( I_k, E_k \in \reals^{3\times3}, \camposition_k\in\reals^3 \) represent the int/extrinsic matrices and positioning vector respectively, of view $k$. Without knowledge of depth from the image plane \(d\), the per-point 
inverse projection in \cref{eq:camera-projection} is under-defined. We argue that depth information is implicitly represented in a scene, and transformer models can learn to create BEV maps, by querying for BEV elements in a key, value sequence, consisting of input features across all views and time instances (see \cref{fig:view-aware-cross-attenion}). In order to help the model develop BEV-to-view feature correspondences, we suggest using a view's int/extrinsic parameters, to positionally encode an input feature $f^{\tau,k}_{i}$ with its view-aware direction embedding, and effectively guide the positional encoding process with camera geometry priors. We do this by creating unit direction vectors $\timeview{\rho} = {E_k}^{-1}{R_k}^{-1}\timeview{x^I} \big|_{d=1}$ for every feature $\timeview{f}$ and its view $k$ pixel coordinates $\timeview{x^I}$ at time index $\tau$.

Instead of using per-feature back-projection to generate BEV features, and following the success of \cite{li2022bevformer,Zhou2022Cross-viewSegmentation,Pan2019Cross-viewSurroundings} we use the function of cross-attention in the transformer decoder \cite{vaswani2017attention} to express the problem as a sequence-to-sequence translation~\cite{Yun2019AreFunctions}. 
\begin{align}
\label{eq:attention}
  \overbrace{\outputsequence}^{\text{output sequence}} 
  &= \CVT (\overbrace{\setindex{q_\text{BEV}}}^{Q} ,  \notag \\
  &\overbrace{\setindex{\timeview{r}}}^K, \notag \\
  &\overbrace{\setindex{\timeview{f}}}^V)  
\end{align}
where the output and query sequence $m_{BEV},Q \in \reals^{n_{\text{BEV}} \times d_m}$ consist of map feature tokens and learnable query tokens, the input sequence consists of features \(\timeview{f} \in \reals^{d_e}\). The output sequence is obtained by quering the direction encoded input sequence with learnable map-queries $Q$.

\[\timeview{r} = \timeview{f} + \phi(\timeview{\rho}) + \iota(\tau)  \]
\[q_i = \Tilde{q}_i + \phi (b_i) - \epsilon(\camposition_k) \]

\[
b_i = E_q^{-1} I_{q}^{-1}\begin{bmatrix}b_{x_i},b_{y_i},1\end{bmatrix}^T , \\ i \in \{0,1,\ldots,h_q\} \times \{0,1,\ldots,w_q\}
\]
where: 
\begin{itemize}
    \item \(\timeview{r}\) are geometry aware feature rays,
    \item $I_q,E_q$ int/extrinsic matrices of a pseudo-camera observing the BEV map, i.e. a pseudocamera with focal length 1, positioned at $(0,0,1)$ and directed towards $-z$,
    \item \(\phi,\epsilon: \reals^3 \rightarrow \reals^{d_e}, \, \iota:\reals \rightarrow \reals^{d_e}\) are linear projection layers, making sure the dimension of the position embeddings match the dimension of the features,
    \item  \(f_i\) corresponds to the feature vector of pixel $x^I_i$ extracted from the associated backbone,
    \item \(q_i \in \reals^{d_m}\) are learnable map-queries,
    \item $d_m, d_e$ are the latent map feature dimensions, and the input feature dimensions respectively.
\end{itemize}

\begin{figure*}[ptb]
	\centering
	\includegraphics[width=0.95\textwidth]{\paperpath/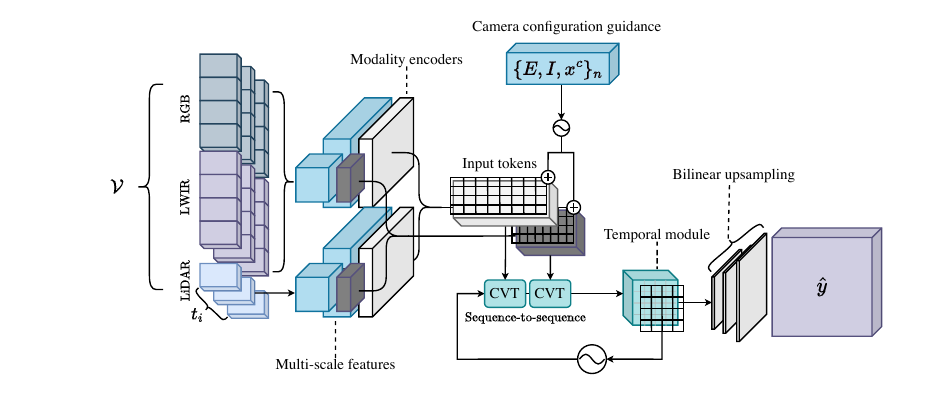}
	\caption{Our proposed fusion framework:  We present a cross-modal and cross-view fusion architecture for the task of BEV segmentation based on camera images and (LiDAR) point clouds. The dense map-queries allow the model to attend to specific regions within each modality, and the multi-scale implementation allows capturing small and large scale context. Temporal information is encoded by 3d convolution blocks. }
	\label{fig:cvt}
\end{figure*}

\textbf{Decoder:}

The semantic segmentation decoder $\mathcal{U}$, consists of three bilinear upsampling layers, that scale the $m_\text{BEV}$ features, to match the latent representation of $y$, i.e.

$$
\hat{y} = \mathcal{U}(m_\text{BEV})
$$

%% file: figures/tex/lidar-channels-pca/lidar-channels-pca_20231219_sample_140_time1657878183.065121.tex
\begingroup%
\makeatletter%
\begin{pgfpicture}%
\pgfpathrectangle{\pgfpointorigin}{\pgfqpoint{2.820262in}{1.752215in}}%
\pgfusepath{use as bounding box, clip}%
\begin{pgfscope}%
\pgfsetbuttcap%
\pgfsetmiterjoin%
\definecolor{currentfill}{rgb}{1.000000,1.000000,1.000000}%
\pgfsetfillcolor{currentfill}%
\pgfsetlinewidth{0.000000pt}%
\definecolor{currentstroke}{rgb}{1.000000,1.000000,1.000000}%
\pgfsetstrokecolor{currentstroke}%
\pgfsetdash{}{0pt}%
\pgfpathmoveto{\pgfqpoint{0.000000in}{0.000000in}}%
\pgfpathlineto{\pgfqpoint{2.820262in}{0.000000in}}%
\pgfpathlineto{\pgfqpoint{2.820262in}{1.752215in}}%
\pgfpathlineto{\pgfqpoint{0.000000in}{1.752215in}}%
\pgfpathlineto{\pgfqpoint{0.000000in}{0.000000in}}%
\pgfpathclose%
\pgfusepath{fill}%
\end{pgfscope}%
\begin{pgfscope}%
\pgfsetbuttcap%
\pgfsetmiterjoin%
\definecolor{currentfill}{rgb}{1.000000,1.000000,1.000000}%
\pgfsetfillcolor{currentfill}%
\pgfsetlinewidth{0.000000pt}%
\definecolor{currentstroke}{rgb}{0.000000,0.000000,0.000000}%
\pgfsetstrokecolor{currentstroke}%
\pgfsetstrokeopacity{0.000000}%
\pgfsetdash{}{0pt}%
\pgfpathmoveto{\pgfqpoint{0.279012in}{0.279012in}}%
\pgfpathlineto{\pgfqpoint{2.720262in}{0.279012in}}%
\pgfpathlineto{\pgfqpoint{2.720262in}{1.652215in}}%
\pgfpathlineto{\pgfqpoint{0.279012in}{1.652215in}}%
\pgfpathlineto{\pgfqpoint{0.279012in}{0.279012in}}%
\pgfpathclose%
\pgfusepath{fill}%
\end{pgfscope}%
\begin{pgfscope}%
\pgfpathrectangle{\pgfqpoint{0.279012in}{0.279012in}}{\pgfqpoint{2.441250in}{1.373203in}}%
\pgfusepath{clip}%
\pgfsys@transformshift{0.279012in}{0.279012in}%
\pgftext[left,bottom]{\includegraphics[interpolate=true,width=2.450000in,height=1.380000in]{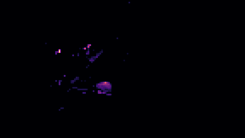}}%
\end{pgfscope}%
\begin{pgfscope}%
\definecolor{textcolor}{rgb}{0.000000,0.000000,0.000000}%
\pgfsetstrokecolor{textcolor}%
\pgfsetfillcolor{textcolor}%
\pgftext[x=1.499637in,y=0.223457in,,top]{\color{textcolor}{\sffamily\fontsize{10.000000}{12.000000}\selectfont\catcode`\^=\active\def^{\ifmmode\sp\else\^{}\fi}\catcode`\%=\active\def
\end{pgfscope}%
\begin{pgfscope}%
\definecolor{textcolor}{rgb}{0.000000,0.000000,0.000000}%
\pgfsetstrokecolor{textcolor}%
\pgfsetfillcolor{textcolor}%
\pgftext[x=0.223457in,y=0.965614in,,bottom,rotate=90.000000]{\color{textcolor}{\sffamily\fontsize{10.000000}{12.000000}\selectfont\catcode`\^=\active\def^{\ifmmode\sp\else\^{}\fi}\catcode`\%=\active\def
\end{pgfscope}%
\begin{pgfscope}%
\pgfsetrectcap%
\pgfsetmiterjoin%
\pgfsetlinewidth{0.803000pt}%
\definecolor{currentstroke}{rgb}{0.000000,0.000000,0.000000}%
\pgfsetstrokecolor{currentstroke}%
\pgfsetdash{}{0pt}%
\pgfpathmoveto{\pgfqpoint{0.279012in}{0.279012in}}%
\pgfpathlineto{\pgfqpoint{0.279012in}{1.652215in}}%
\pgfusepath{stroke}%
\end{pgfscope}%
\begin{pgfscope}%
\pgfsetrectcap%
\pgfsetmiterjoin%
\pgfsetlinewidth{0.803000pt}%
\definecolor{currentstroke}{rgb}{0.000000,0.000000,0.000000}%
\pgfsetstrokecolor{currentstroke}%
\pgfsetdash{}{0pt}%
\pgfpathmoveto{\pgfqpoint{2.720262in}{0.279012in}}%
\pgfpathlineto{\pgfqpoint{2.720262in}{1.652215in}}%
\pgfusepath{stroke}%
\end{pgfscope}%
\begin{pgfscope}%
\pgfsetrectcap%
\pgfsetmiterjoin%
\pgfsetlinewidth{0.803000pt}%
\definecolor{currentstroke}{rgb}{0.000000,0.000000,0.000000}%
\pgfsetstrokecolor{currentstroke}%
\pgfsetdash{}{0pt}%
\pgfpathmoveto{\pgfqpoint{0.279012in}{0.279012in}}%
\pgfpathlineto{\pgfqpoint{2.720262in}{0.279012in}}%
\pgfusepath{stroke}%
\end{pgfscope}%
\begin{pgfscope}%
\pgfsetrectcap%
\pgfsetmiterjoin%
\pgfsetlinewidth{0.803000pt}%
\definecolor{currentstroke}{rgb}{0.000000,0.000000,0.000000}%
\pgfsetstrokecolor{currentstroke}%
\pgfsetdash{}{0pt}%
\pgfpathmoveto{\pgfqpoint{0.279012in}{1.652215in}}%
\pgfpathlineto{\pgfqpoint{2.720262in}{1.652215in}}%
\pgfusepath{stroke}%
\end{pgfscope}%
\end{pgfpicture}%
\makeatother%
\endgroup%

%% file: chapters/proposed_approach/multimodality.tex
\subsection{Motivation for Multi-modality}\label{subseq:multimodality}
The principle of multi-modality posits that the integration of additional sensors, in conjunction with traditional RGB cameras, can potentially enhance the model's ability to reason the depth of the ship's surroundings and the category to which they belong. Regarding depth, the infrared and point cloud data inherently contain that information. Focusing on the segmentation aspect, the lidar and radar data provide localization information that can not be found in camera images. As a result, we hypothetize that downstream task performance can be improved given their effective deep fusion.

\begin{figure}[ptb]
	\centering
	\includegraphics[width=0.9\columnwidth]{\paperpath/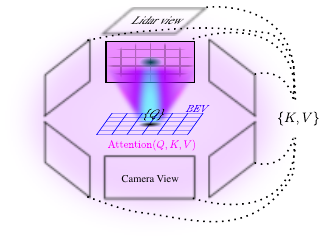}
	\caption{View-aware cross attention. Pixel positions are lifted to unit direction vector coordinates, which are appended to the features as positional embeddings.}
	\label{fig:view-aware-cross-attenion}
\end{figure}

%% file: chapters/proposed_approach/temporality.tex
\subsection{Temporal features}\label{subseq:temporality}
By integrating historical BEV information with current environmental data, models can stabilize perception results, represent temporarily occluded objects, accumulate observations for map generation, and improve the detection of moving or heavily occluded objects \cite{10.48550/arxiv.2207.08536,liu2023bevfusion}. In our work, four out of five classes are static (buoys, land, shoreline, water). The displacement of these classes between consecutive images and BEV maps therefore, depends exclusively on the own-ship's motion. The ship's pose is, in most cases,  measurable through GNSS.

The \emph{target} class, however, might display non-linear motion dynamics, and the displacement of this class's instances between consecutive images and BEV maps, depends on the combined relative displacement between the moving targets and own-ship. The \emph{target} instances motion dynamics are therefore relatively unpredictable, especially without being able to measure their velocity. We hypothesize that by allowing a model to simultaneously process temporal information across multiple time instances, it can learn object motion dynamics and avoid occlusion, as demonstrated in \cite{li2022bevformer}.

We extend the work of \cite{Zhou2022Cross-viewSegmentation} and incorporate a temporal aggregation module, acting on heterogeneous multi-view features. We attempt this by using ego-motion to align the output of the cross-attention layers across different time instants, and then extracting spatio-temporal BEV features, with 3D convolution kernels.

\textbf{Temporal Alignment}
To align the BEV feature maps from consecutive time instances, similarly to \cite{hu2021fiery,Zhang2022BEVerse:Driving,huang2021bevdet}, we use ego-motion pose differences, to align the feature maps between past/future time instances.


\textbf{3D Convolutional Layers}
The output of the alignment process are $t$ aligned BEV feature maps. We then use a 3D convolution layer without padding, to extract spatio-temporal BEV features.

%% file: chapters/evaluation.tex
\section{Experiments}\label{sec:Eval}

\subsection{Implementation details}

\textbf{Model parameters:} 
We use transfer learning and EfficientNet \cite{tan2019efficientnet} checkpoints, to implement the common RGB and LWIR camera feature extractor. For the LWIR images, which are single-channel, we employ channel replication to create a three-channel input compatible with the RGB-based network architecture. We standardize all images to 224x480 resolution and similar to \cite{Zhou2022Cross-viewSegmentation}, we extract features in two 8x and 16x down-scaled resolutions, i.e., (28,60) and (14,30) and $d_e =128$ channels.

For the learnable BEV queries, we choose $n_{\text{BEV}}=25\times25=625$ and $d_m=128$. For the multi-head cross-attention blocks, we choose 4 heads and inner dimension 64. The decoder consist of three bi-linear upsampling and convolution layers, upscaling the latent map feature tokens $m_{\text{BEV}}$ to the prediction outputs $\hat{y} \in \reals^{200\times200\times5}$, which corresponds to $3 \text{ m/pixel}$ orthographic projection maps, one for each class, covering 600m in width and height (see also \cref{fig:prediction-heatmap}).

For the temporal aggregation module, we choose $t=3$ time instances in \cref{eq:heterogeneous-data} and select samples with $\SI{5}{\second}$ difference between them.

\textbf{Augmentations:} 
BEV Perception needs to remain robust to geometrical perturbations in images. Such perturbations are either non-constant, for example as consequence of the vessel's floating movements, or static, i.e. due to imprecise calibration parameters. The goal of train-time augmentations is to make the model invariant to such perturbations, but also regularize the training process, by reducing over-fitting. Train-time data augmentation plays a vital role in improving BEV mapping performance, with \cite{yang2023bevformer} reporting substantial benefits of train-time augmentations in their ablation studies. 

We apply random geometric augmentation to the sensor views, implemented with random cropping and rotation, where we update the int/extrinsic matrices correspondingly \cite{harley2023simple}. Following the geometric transformations, we apply random color variation augmentations, trying to cover different light exposure conditions, as well as random camera dropout augmentation, where one or more of the sensors are being randomly blacked out during training.


\textbf{Training}:
We train on 2 NVidia A100 GPUS for 120 epochs, over 8 hours with batch size of 12 samples, using binary focal loss. We optimize using AdamW \cite{loshchilov2017decoupled} optimizer with cyclical learning rate scheduling \cite{smith2019super} and a maximum learning rate of $4\times10^-3$.

\textbf{Evaluation:}
We apply softmax and argmax to the predicted BEV maps $\hat{y}$ across the channel dimension and calculate the multi-class \gls{IOU}. We repeat the evaluation over a multitude of experiments, and verify the performance benefits of modal and temporal fusion. The evaluation is summarized in \cref{tab:experiment-results}. Furthermore, we empirically evaluate the predictive performance of the model by looking at individual class prediction heat-maps in \cref{fig:prediction-heatmap}, but also by back-tracing salience heta-maps from the cross-attention similarity matrices in the transformer blocks (\cref{fig:heads0classtargets}).

\input{\paperpath/chapters/tables/results}

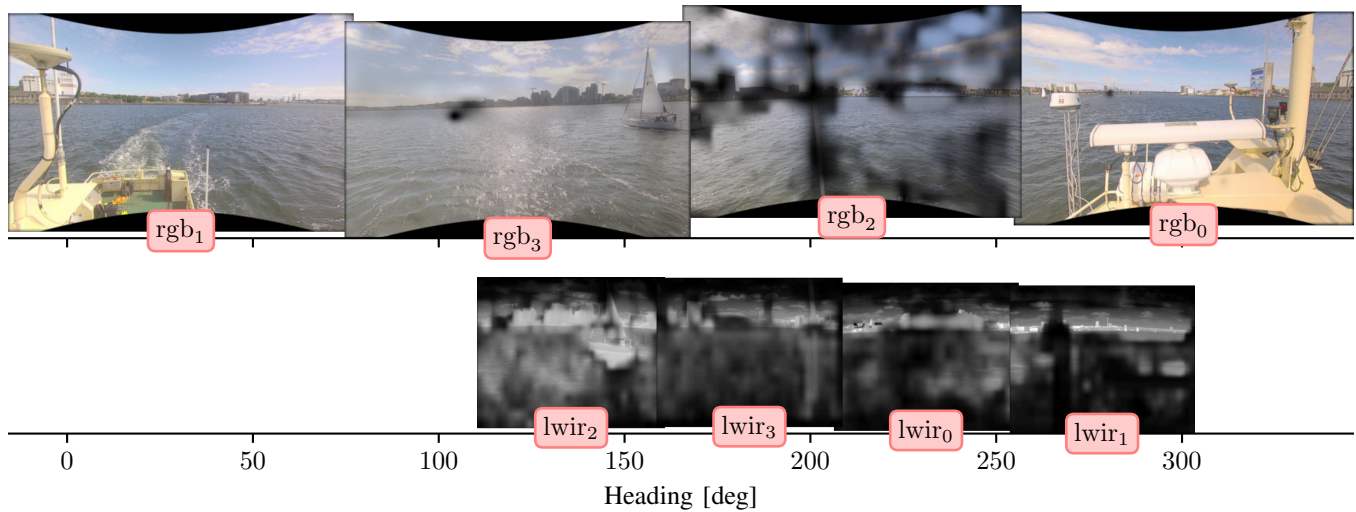
\begin{figure*}[ht]
    \centering
    \resizebox{1.0\linewidth}{!}{
    \import{\paperpath/figures/tex/camera-layout-aligned-with-attention}{camera-layout-aligned-with-attention_1657876089.234451.tex}
    }
    \caption{Overlaying attention map on the input data. The figure illustrates the correlation between BEV and camera view representations. Image position on each row adjusted for camera $k$ extrinsic parameters $I_k$ (see also \cref{fig:heads0classtargets}).}
    \label{fig:camear-layout-attention-map}
\end{figure*}

\subsection{Explainability}
\begin{figure*}[ht]
\centering
\includegraphics[width=\textwidth]{\paperpath/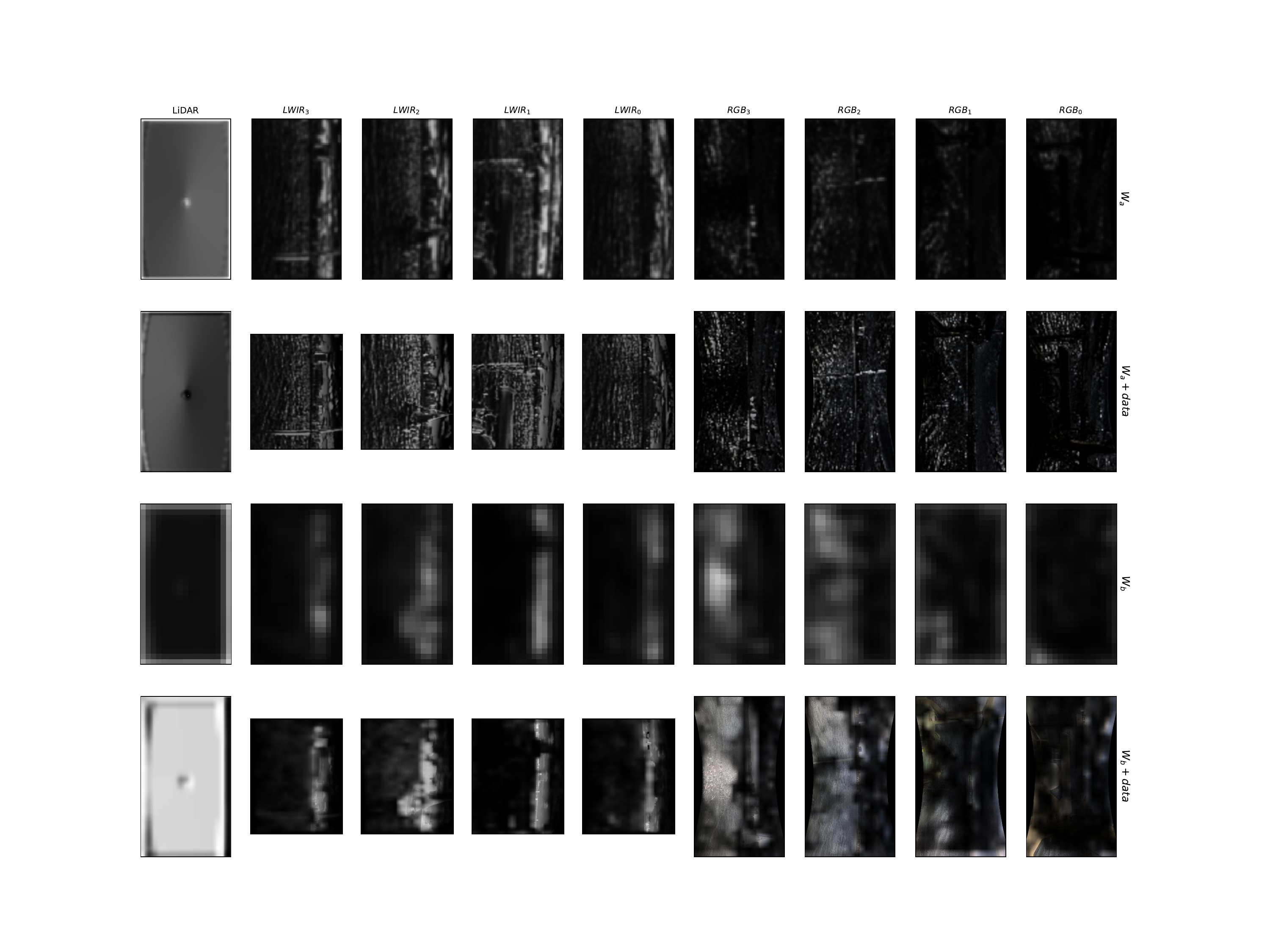}
\caption{Attention maps $W_{\text{probe}}$ calculated from the two cross-attention modules, that operate on different input feature resolutions (\cref{fig:cvt}). The produced score map demonstrate the ability of the model to focus on salient regions that correspond to a BEV region (here to the regions highlighted by red color in \cref{fig:target-features}) discarding the non-relevant
areas in feature fusion.}
\label{fig:heads0classtargets}
\end{figure*}

We use salience maps, to demonstrate the model's ability to focus on specific regions within the views $\Vset$. \Cref{fig:heads0classtargets} illustrates the use of extracted attention maps within the model's cross attention blocks to generate saliency maps and overlay them on the camera views. The illustration demonstrates that the model is able to attest to significant parts of the input sequence $\Vset$ during task inference. The attention mechanism allows the model to assign different weights to different parts of the data, providing insight into which areas influence the model's predictions the most. By revisiting \cref{eq:attention}, the look-up function is implemented by
\[ \text{Attention}(\mathbf{Q}, \mathbf{K}, \mathbf{V}) = \mathbf{W} \mathbf{V} \]

\[ \mathbf{W} = \text{softmax}\left(\frac{\mathbf{Q}\mathbf{K}^T}{\sqrt{d_k}}\right) \in \reals^{n_h \times (tVN) \times N_B} \]

where \(n_h\) represents the number of heads in Multi-head attention.

\newcommand{\nprobe}{n_{\text{probe}}}
To visualise the attention heat-maps, we select a region of interest in the produced BEV maps, that corresponds to a set of indices $\nprobe < N_B$ such that \(\mathbf{W}_{\text{probe}} \subset \mathbf{W}\). The attention matrix $\mathbf{W}_{\text{probe}}$ is indexed on current time instance, while information along the number of heads \(n_h\) is reduced by projecting $\mathbf{W}$ on its first principal component along the head dimension. The attention maps are bi-linearly up-sampled to match the input camera view resolution, resulting in attention matrices \(\mathbf{W}^v_{\text{probe}} \in \reals^{H\times W}\), for \(v=\{1,\ldots, V\}\) that can be visualised on top camera views, as in \cref{fig:heads0classtargets}.

%% file: chapters/tables/results.tex
\begin{table*}[ht]
    \centering
    \caption{Results in terms of mean Intersection over Union (mIOU).}
    \label{tab:experiment-results}
    \begin{tabular}{lllllll}
    \toprule
    Method                & Modality               & Boat   & Buoy   & Water   & Shoreline   & Land   \\ 
    \midrule
    Standard              & RGB                    & 11\%   & 30\%   & 70\%    & 22\%        & 92\%   \\ 
    Standard              & RGB, LWIR              & 13\%   & 32\%   & 78\%    & 23\%        & 93\%   \\ 
    Standard              & RGB, LWIR, LiDAR       & 15\%   & 34\%   & 80\%    & 26\%        & 95\%   \\   
    Temporal              & RGB, LWIR, LiDAR       & 17\%   & 36\%   & 85\%    & 27\%        & 96\%   \\   
    Temporal w. alignment & RGB, LWIR, LiDAR       & 17\%   & 37\%   & 91\%    & 31\%        & 96\%   \\
    \bottomrule
    \end{tabular}
    \end{table*}

%% file: figures/tex/camera-layout-aligned-with-attention/camera-layout-aligned-with-attention_1657876089.234451.tex
\begingroup%
\makeatletter%
\begin{pgfpicture}%
\pgfpathrectangle{\pgfpointorigin}{\pgfqpoint{7.079625in}{2.667682in}}%
\pgfusepath{use as bounding box, clip}%
\begin{pgfscope}%
\pgfsetbuttcap%
\pgfsetmiterjoin%
\definecolor{currentfill}{rgb}{1.000000,1.000000,1.000000}%
\pgfsetfillcolor{currentfill}%
\pgfsetlinewidth{0.000000pt}%
\definecolor{currentstroke}{rgb}{1.000000,1.000000,1.000000}%
\pgfsetstrokecolor{currentstroke}%
\pgfsetdash{}{0pt}%
\pgfpathmoveto{\pgfqpoint{0.000000in}{0.000000in}}%
\pgfpathlineto{\pgfqpoint{7.079625in}{0.000000in}}%
\pgfpathlineto{\pgfqpoint{7.079625in}{2.667682in}}%
\pgfpathlineto{\pgfqpoint{0.000000in}{2.667682in}}%
\pgfpathlineto{\pgfqpoint{0.000000in}{0.000000in}}%
\pgfpathclose%
\pgfusepath{fill}%
\end{pgfscope}%
\begin{pgfscope}%
\pgfsetbuttcap%
\pgfsetmiterjoin%
\definecolor{currentfill}{rgb}{1.000000,1.000000,1.000000}%
\pgfsetfillcolor{currentfill}%
\pgfsetlinewidth{0.000000pt}%
\definecolor{currentstroke}{rgb}{0.000000,0.000000,0.000000}%
\pgfsetstrokecolor{currentstroke}%
\pgfsetstrokeopacity{0.000000}%
\pgfsetdash{}{0pt}%
\pgfpathmoveto{\pgfqpoint{0.000000in}{1.443745in}}%
\pgfpathlineto{\pgfqpoint{7.079625in}{1.443745in}}%
\pgfpathlineto{\pgfqpoint{7.079625in}{2.667682in}}%
\pgfpathlineto{\pgfqpoint{0.000000in}{2.667682in}}%
\pgfpathlineto{\pgfqpoint{0.000000in}{1.443745in}}%
\pgfpathclose%
\pgfusepath{fill}%
\end{pgfscope}%
\begin{pgfscope}%
\pgfpathrectangle{\pgfqpoint{0.000000in}{1.443745in}}{\pgfqpoint{7.079625in}{1.223937in}}%
\pgfusepath{clip}%
\pgfsys@transformshift{0.000000in}{1.444000in}%
\pgftext[left,bottom]{\includegraphics[interpolate=true,width=7.080000in,height=1.224000in]{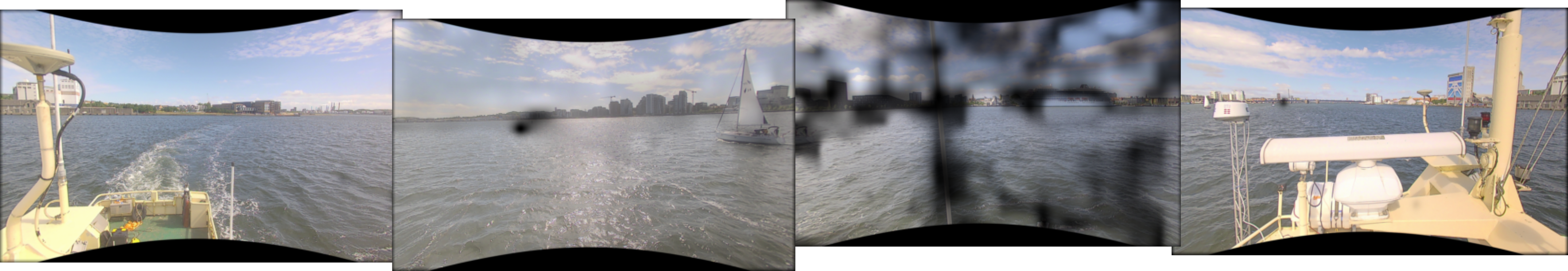}}%
\end{pgfscope}%
\begin{pgfscope}%
\pgfsetbuttcap%
\pgfsetroundjoin%
\definecolor{currentfill}{rgb}{0.000000,0.000000,0.000000}%
\pgfsetfillcolor{currentfill}%
\pgfsetlinewidth{0.803000pt}%
\definecolor{currentstroke}{rgb}{0.000000,0.000000,0.000000}%
\pgfsetstrokecolor{currentstroke}%
\pgfsetdash{}{0pt}%
\pgfsys@defobject{currentmarker}{\pgfqpoint{0.000000in}{-0.048611in}}{\pgfqpoint{0.000000in}{0.000000in}}{%
\pgfpathmoveto{\pgfqpoint{0.000000in}{0.000000in}}%
\pgfpathlineto{\pgfqpoint{0.000000in}{-0.048611in}}%
\pgfusepath{stroke,fill}%
}%
\begin{pgfscope}%
\pgfsys@transformshift{0.310217in}{1.443745in}%
\pgfsys@useobject{currentmarker}{}%
\end{pgfscope}%
\end{pgfscope}%
\begin{pgfscope}%
\pgfsetbuttcap%
\pgfsetroundjoin%
\definecolor{currentfill}{rgb}{0.000000,0.000000,0.000000}%
\pgfsetfillcolor{currentfill}%
\pgfsetlinewidth{0.803000pt}%
\definecolor{currentstroke}{rgb}{0.000000,0.000000,0.000000}%
\pgfsetstrokecolor{currentstroke}%
\pgfsetdash{}{0pt}%
\pgfsys@defobject{currentmarker}{\pgfqpoint{0.000000in}{-0.048611in}}{\pgfqpoint{0.000000in}{0.000000in}}{%
\pgfpathmoveto{\pgfqpoint{0.000000in}{0.000000in}}%
\pgfpathlineto{\pgfqpoint{0.000000in}{-0.048611in}}%
\pgfusepath{stroke,fill}%
}%
\begin{pgfscope}%
\pgfsys@transformshift{1.287750in}{1.443745in}%
\pgfsys@useobject{currentmarker}{}%
\end{pgfscope}%
\end{pgfscope}%
\begin{pgfscope}%
\pgfsetbuttcap%
\pgfsetroundjoin%
\definecolor{currentfill}{rgb}{0.000000,0.000000,0.000000}%
\pgfsetfillcolor{currentfill}%
\pgfsetlinewidth{0.803000pt}%
\definecolor{currentstroke}{rgb}{0.000000,0.000000,0.000000}%
\pgfsetstrokecolor{currentstroke}%
\pgfsetdash{}{0pt}%
\pgfsys@defobject{currentmarker}{\pgfqpoint{0.000000in}{-0.048611in}}{\pgfqpoint{0.000000in}{0.000000in}}{%
\pgfpathmoveto{\pgfqpoint{0.000000in}{0.000000in}}%
\pgfpathlineto{\pgfqpoint{0.000000in}{-0.048611in}}%
\pgfusepath{stroke,fill}%
}%
\begin{pgfscope}%
\pgfsys@transformshift{2.265283in}{1.443745in}%
\pgfsys@useobject{currentmarker}{}%
\end{pgfscope}%
\end{pgfscope}%
\begin{pgfscope}%
\pgfsetbuttcap%
\pgfsetroundjoin%
\definecolor{currentfill}{rgb}{0.000000,0.000000,0.000000}%
\pgfsetfillcolor{currentfill}%
\pgfsetlinewidth{0.803000pt}%
\definecolor{currentstroke}{rgb}{0.000000,0.000000,0.000000}%
\pgfsetstrokecolor{currentstroke}%
\pgfsetdash{}{0pt}%
\pgfsys@defobject{currentmarker}{\pgfqpoint{0.000000in}{-0.048611in}}{\pgfqpoint{0.000000in}{0.000000in}}{%
\pgfpathmoveto{\pgfqpoint{0.000000in}{0.000000in}}%
\pgfpathlineto{\pgfqpoint{0.000000in}{-0.048611in}}%
\pgfusepath{stroke,fill}%
}%
\begin{pgfscope}%
\pgfsys@transformshift{3.242816in}{1.443745in}%
\pgfsys@useobject{currentmarker}{}%
\end{pgfscope}%
\end{pgfscope}%
\begin{pgfscope}%
\pgfsetbuttcap%
\pgfsetroundjoin%
\definecolor{currentfill}{rgb}{0.000000,0.000000,0.000000}%
\pgfsetfillcolor{currentfill}%
\pgfsetlinewidth{0.803000pt}%
\definecolor{currentstroke}{rgb}{0.000000,0.000000,0.000000}%
\pgfsetstrokecolor{currentstroke}%
\pgfsetdash{}{0pt}%
\pgfsys@defobject{currentmarker}{\pgfqpoint{0.000000in}{-0.048611in}}{\pgfqpoint{0.000000in}{0.000000in}}{%
\pgfpathmoveto{\pgfqpoint{0.000000in}{0.000000in}}%
\pgfpathlineto{\pgfqpoint{0.000000in}{-0.048611in}}%
\pgfusepath{stroke,fill}%
}%
\begin{pgfscope}%
\pgfsys@transformshift{4.220348in}{1.443745in}%
\pgfsys@useobject{currentmarker}{}%
\end{pgfscope}%
\end{pgfscope}%
\begin{pgfscope}%
\pgfsetbuttcap%
\pgfsetroundjoin%
\definecolor{currentfill}{rgb}{0.000000,0.000000,0.000000}%
\pgfsetfillcolor{currentfill}%
\pgfsetlinewidth{0.803000pt}%
\definecolor{currentstroke}{rgb}{0.000000,0.000000,0.000000}%
\pgfsetstrokecolor{currentstroke}%
\pgfsetdash{}{0pt}%
\pgfsys@defobject{currentmarker}{\pgfqpoint{0.000000in}{-0.048611in}}{\pgfqpoint{0.000000in}{0.000000in}}{%
\pgfpathmoveto{\pgfqpoint{0.000000in}{0.000000in}}%
\pgfpathlineto{\pgfqpoint{0.000000in}{-0.048611in}}%
\pgfusepath{stroke,fill}%
}%
\begin{pgfscope}%
\pgfsys@transformshift{5.197881in}{1.443745in}%
\pgfsys@useobject{currentmarker}{}%
\end{pgfscope}%
\end{pgfscope}%
\begin{pgfscope}%
\pgfsetbuttcap%
\pgfsetroundjoin%
\definecolor{currentfill}{rgb}{0.000000,0.000000,0.000000}%
\pgfsetfillcolor{currentfill}%
\pgfsetlinewidth{0.803000pt}%
\definecolor{currentstroke}{rgb}{0.000000,0.000000,0.000000}%
\pgfsetstrokecolor{currentstroke}%
\pgfsetdash{}{0pt}%
\pgfsys@defobject{currentmarker}{\pgfqpoint{0.000000in}{-0.048611in}}{\pgfqpoint{0.000000in}{0.000000in}}{%
\pgfpathmoveto{\pgfqpoint{0.000000in}{0.000000in}}%
\pgfpathlineto{\pgfqpoint{0.000000in}{-0.048611in}}%
\pgfusepath{stroke,fill}%
}%
\begin{pgfscope}%
\pgfsys@transformshift{6.175414in}{1.443745in}%
\pgfsys@useobject{currentmarker}{}%
\end{pgfscope}%
\end{pgfscope}%
\begin{pgfscope}%
\pgfsetrectcap%
\pgfsetmiterjoin%
\pgfsetlinewidth{0.803000pt}%
\definecolor{currentstroke}{rgb}{0.000000,0.000000,0.000000}%
\pgfsetstrokecolor{currentstroke}%
\pgfsetdash{}{0pt}%
\pgfpathmoveto{\pgfqpoint{0.000000in}{1.443745in}}%
\pgfpathlineto{\pgfqpoint{7.079625in}{1.443745in}}%
\pgfusepath{stroke}%
\end{pgfscope}%
\begin{pgfscope}%
\pgfsetbuttcap%
\pgfsetmiterjoin%
\definecolor{currentfill}{rgb}{1.000000,0.800000,0.800000}%
\pgfsetfillcolor{currentfill}%
\pgfsetlinewidth{1.003750pt}%
\definecolor{currentstroke}{rgb}{1.000000,0.500000,0.500000}%
\pgfsetstrokecolor{currentstroke}%
\pgfsetdash{}{0pt}%
\pgfpathmoveto{\pgfqpoint{6.053251in}{1.408238in}}%
\pgfpathlineto{\pgfqpoint{6.316562in}{1.408238in}}%
\pgfpathquadraticcurveto{\pgfqpoint{6.358228in}{1.408238in}}{\pgfqpoint{6.358228in}{1.449905in}}%
\pgfpathlineto{\pgfqpoint{6.358228in}{1.580306in}}%
\pgfpathquadraticcurveto{\pgfqpoint{6.358228in}{1.621972in}}{\pgfqpoint{6.316562in}{1.621972in}}%
\pgfpathlineto{\pgfqpoint{6.053251in}{1.621972in}}%
\pgfpathquadraticcurveto{\pgfqpoint{6.011584in}{1.621972in}}{\pgfqpoint{6.011584in}{1.580306in}}%
\pgfpathlineto{\pgfqpoint{6.011584in}{1.449905in}}%
\pgfpathquadraticcurveto{\pgfqpoint{6.011584in}{1.408238in}}{\pgfqpoint{6.053251in}{1.408238in}}%
\pgfpathlineto{\pgfqpoint{6.053251in}{1.408238in}}%
\pgfpathclose%
\pgfusepath{stroke,fill}%
\end{pgfscope}%
\begin{pgfscope}%
\definecolor{textcolor}{rgb}{0.000000,0.000000,0.000000}%
\pgfsetstrokecolor{textcolor}%
\pgfsetfillcolor{textcolor}%
\pgftext[x=6.184906in,y=1.515105in,,]{\color{textcolor}{\rmfamily\fontsize{10.000000}{12.000000}\selectfont\catcode`\^=\active\def^{\ifmmode\sp\else\^{}\fi}\catcode`\%=\active\def
\end{pgfscope}%
\begin{pgfscope}%
\pgfsetbuttcap%
\pgfsetmiterjoin%
\definecolor{currentfill}{rgb}{1.000000,0.800000,0.800000}%
\pgfsetfillcolor{currentfill}%
\pgfsetlinewidth{1.003750pt}%
\definecolor{currentstroke}{rgb}{1.000000,0.500000,0.500000}%
\pgfsetstrokecolor{currentstroke}%
\pgfsetdash{}{0pt}%
\pgfpathmoveto{\pgfqpoint{0.777256in}{1.375450in}}%
\pgfpathlineto{\pgfqpoint{1.040567in}{1.375450in}}%
\pgfpathquadraticcurveto{\pgfqpoint{1.082234in}{1.375450in}}{\pgfqpoint{1.082234in}{1.417117in}}%
\pgfpathlineto{\pgfqpoint{1.082234in}{1.547518in}}%
\pgfpathquadraticcurveto{\pgfqpoint{1.082234in}{1.589184in}}{\pgfqpoint{1.040567in}{1.589184in}}%
\pgfpathlineto{\pgfqpoint{0.777256in}{1.589184in}}%
\pgfpathquadraticcurveto{\pgfqpoint{0.735590in}{1.589184in}}{\pgfqpoint{0.735590in}{1.547518in}}%
\pgfpathlineto{\pgfqpoint{0.735590in}{1.417117in}}%
\pgfpathquadraticcurveto{\pgfqpoint{0.735590in}{1.375450in}}{\pgfqpoint{0.777256in}{1.375450in}}%
\pgfpathlineto{\pgfqpoint{0.777256in}{1.375450in}}%
\pgfpathclose%
\pgfusepath{stroke,fill}%
\end{pgfscope}%
\begin{pgfscope}%
\definecolor{textcolor}{rgb}{0.000000,0.000000,0.000000}%
\pgfsetstrokecolor{textcolor}%
\pgfsetfillcolor{textcolor}%
\pgftext[x=0.908912in,y=1.482317in,,]{\color{textcolor}{\rmfamily\fontsize{10.000000}{12.000000}\selectfont\catcode`\^=\active\def^{\ifmmode\sp\else\^{}\fi}\catcode`\%=\active\def
\end{pgfscope}%
\begin{pgfscope}%
\pgfsetbuttcap%
\pgfsetmiterjoin%
\definecolor{currentfill}{rgb}{1.000000,0.800000,0.800000}%
\pgfsetfillcolor{currentfill}%
\pgfsetlinewidth{1.003750pt}%
\definecolor{currentstroke}{rgb}{1.000000,0.500000,0.500000}%
\pgfsetstrokecolor{currentstroke}%
\pgfsetdash{}{0pt}%
\pgfpathmoveto{\pgfqpoint{4.307364in}{1.447343in}}%
\pgfpathlineto{\pgfqpoint{4.570675in}{1.447343in}}%
\pgfpathquadraticcurveto{\pgfqpoint{4.612342in}{1.447343in}}{\pgfqpoint{4.612342in}{1.489010in}}%
\pgfpathlineto{\pgfqpoint{4.612342in}{1.619411in}}%
\pgfpathquadraticcurveto{\pgfqpoint{4.612342in}{1.661077in}}{\pgfqpoint{4.570675in}{1.661077in}}%
\pgfpathlineto{\pgfqpoint{4.307364in}{1.661077in}}%
\pgfpathquadraticcurveto{\pgfqpoint{4.265698in}{1.661077in}}{\pgfqpoint{4.265698in}{1.619411in}}%
\pgfpathlineto{\pgfqpoint{4.265698in}{1.489010in}}%
\pgfpathquadraticcurveto{\pgfqpoint{4.265698in}{1.447343in}}{\pgfqpoint{4.307364in}{1.447343in}}%
\pgfpathlineto{\pgfqpoint{4.307364in}{1.447343in}}%
\pgfpathclose%
\pgfusepath{stroke,fill}%
\end{pgfscope}%
\begin{pgfscope}%
\definecolor{textcolor}{rgb}{0.000000,0.000000,0.000000}%
\pgfsetstrokecolor{textcolor}%
\pgfsetfillcolor{textcolor}%
\pgftext[x=4.439020in,y=1.554210in,,]{\color{textcolor}{\rmfamily\fontsize{10.000000}{12.000000}\selectfont\catcode`\^=\active\def^{\ifmmode\sp\else\^{}\fi}\catcode`\%=\active\def
\end{pgfscope}%
\begin{pgfscope}%
\pgfsetbuttcap%
\pgfsetmiterjoin%
\definecolor{currentfill}{rgb}{1.000000,0.800000,0.800000}%
\pgfsetfillcolor{currentfill}%
\pgfsetlinewidth{1.003750pt}%
\definecolor{currentstroke}{rgb}{1.000000,0.500000,0.500000}%
\pgfsetstrokecolor{currentstroke}%
\pgfsetdash{}{0pt}%
\pgfpathmoveto{\pgfqpoint{2.549282in}{1.336878in}}%
\pgfpathlineto{\pgfqpoint{2.812593in}{1.336878in}}%
\pgfpathquadraticcurveto{\pgfqpoint{2.854259in}{1.336878in}}{\pgfqpoint{2.854259in}{1.378544in}}%
\pgfpathlineto{\pgfqpoint{2.854259in}{1.508945in}}%
\pgfpathquadraticcurveto{\pgfqpoint{2.854259in}{1.550612in}}{\pgfqpoint{2.812593in}{1.550612in}}%
\pgfpathlineto{\pgfqpoint{2.549282in}{1.550612in}}%
\pgfpathquadraticcurveto{\pgfqpoint{2.507615in}{1.550612in}}{\pgfqpoint{2.507615in}{1.508945in}}%
\pgfpathlineto{\pgfqpoint{2.507615in}{1.378544in}}%
\pgfpathquadraticcurveto{\pgfqpoint{2.507615in}{1.336878in}}{\pgfqpoint{2.549282in}{1.336878in}}%
\pgfpathlineto{\pgfqpoint{2.549282in}{1.336878in}}%
\pgfpathclose%
\pgfusepath{stroke,fill}%
\end{pgfscope}%
\begin{pgfscope}%
\definecolor{textcolor}{rgb}{0.000000,0.000000,0.000000}%
\pgfsetstrokecolor{textcolor}%
\pgfsetfillcolor{textcolor}%
\pgftext[x=2.680937in,y=1.443745in,,]{\color{textcolor}{\rmfamily\fontsize{10.000000}{12.000000}\selectfont\catcode`\^=\active\def^{\ifmmode\sp\else\^{}\fi}\catcode`\%=\active\def
\end{pgfscope}%
\begin{pgfscope}%
\pgfsetbuttcap%
\pgfsetmiterjoin%
\definecolor{currentfill}{rgb}{1.000000,1.000000,1.000000}%
\pgfsetfillcolor{currentfill}%
\pgfsetlinewidth{0.000000pt}%
\definecolor{currentstroke}{rgb}{0.000000,0.000000,0.000000}%
\pgfsetstrokecolor{currentstroke}%
\pgfsetstrokeopacity{0.000000}%
\pgfsetdash{}{0pt}%
\pgfpathmoveto{\pgfqpoint{0.000000in}{0.415123in}}%
\pgfpathlineto{\pgfqpoint{7.079625in}{0.415123in}}%
\pgfpathlineto{\pgfqpoint{7.079625in}{1.234421in}}%
\pgfpathlineto{\pgfqpoint{0.000000in}{1.234421in}}%
\pgfpathlineto{\pgfqpoint{0.000000in}{0.415123in}}%
\pgfpathclose%
\pgfusepath{fill}%
\end{pgfscope}%
\begin{pgfscope}%
\pgfpathrectangle{\pgfqpoint{0.000000in}{0.415123in}}{\pgfqpoint{7.079625in}{0.819298in}}%
\pgfusepath{clip}%
\pgfsys@transformshift{2.468000in}{0.416000in}%
\pgftext[left,bottom]{\includegraphics[interpolate=true,width=3.774000in,height=0.820000in]{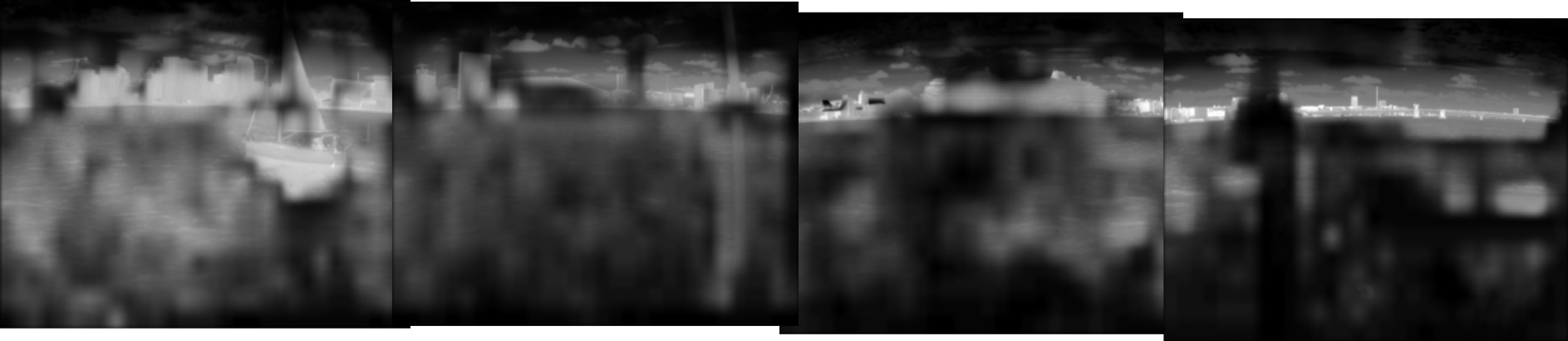}}%
\end{pgfscope}%
\begin{pgfscope}%
\pgfsetbuttcap%
\pgfsetroundjoin%
\definecolor{currentfill}{rgb}{0.000000,0.000000,0.000000}%
\pgfsetfillcolor{currentfill}%
\pgfsetlinewidth{0.803000pt}%
\definecolor{currentstroke}{rgb}{0.000000,0.000000,0.000000}%
\pgfsetstrokecolor{currentstroke}%
\pgfsetdash{}{0pt}%
\pgfsys@defobject{currentmarker}{\pgfqpoint{0.000000in}{-0.048611in}}{\pgfqpoint{0.000000in}{0.000000in}}{%
\pgfpathmoveto{\pgfqpoint{0.000000in}{0.000000in}}%
\pgfpathlineto{\pgfqpoint{0.000000in}{-0.048611in}}%
\pgfusepath{stroke,fill}%
}%
\begin{pgfscope}%
\pgfsys@transformshift{0.310217in}{0.415123in}%
\pgfsys@useobject{currentmarker}{}%
\end{pgfscope}%
\end{pgfscope}%
\begin{pgfscope}%
\definecolor{textcolor}{rgb}{0.000000,0.000000,0.000000}%
\pgfsetstrokecolor{textcolor}%
\pgfsetfillcolor{textcolor}%
\pgftext[x=0.310217in,y=0.317901in,,top]{\color{textcolor}{\rmfamily\fontsize{10.000000}{12.000000}\selectfont\catcode`\^=\active\def^{\ifmmode\sp\else\^{}\fi}\catcode`\%=\active\def
\end{pgfscope}%
\begin{pgfscope}%
\pgfsetbuttcap%
\pgfsetroundjoin%
\definecolor{currentfill}{rgb}{0.000000,0.000000,0.000000}%
\pgfsetfillcolor{currentfill}%
\pgfsetlinewidth{0.803000pt}%
\definecolor{currentstroke}{rgb}{0.000000,0.000000,0.000000}%
\pgfsetstrokecolor{currentstroke}%
\pgfsetdash{}{0pt}%
\pgfsys@defobject{currentmarker}{\pgfqpoint{0.000000in}{-0.048611in}}{\pgfqpoint{0.000000in}{0.000000in}}{%
\pgfpathmoveto{\pgfqpoint{0.000000in}{0.000000in}}%
\pgfpathlineto{\pgfqpoint{0.000000in}{-0.048611in}}%
\pgfusepath{stroke,fill}%
}%
\begin{pgfscope}%
\pgfsys@transformshift{1.287750in}{0.415123in}%
\pgfsys@useobject{currentmarker}{}%
\end{pgfscope}%
\end{pgfscope}%
\begin{pgfscope}%
\definecolor{textcolor}{rgb}{0.000000,0.000000,0.000000}%
\pgfsetstrokecolor{textcolor}%
\pgfsetfillcolor{textcolor}%
\pgftext[x=1.287750in,y=0.317901in,,top]{\color{textcolor}{\rmfamily\fontsize{10.000000}{12.000000}\selectfont\catcode`\^=\active\def^{\ifmmode\sp\else\^{}\fi}\catcode`\%=\active\def
\end{pgfscope}%
\begin{pgfscope}%
\pgfsetbuttcap%
\pgfsetroundjoin%
\definecolor{currentfill}{rgb}{0.000000,0.000000,0.000000}%
\pgfsetfillcolor{currentfill}%
\pgfsetlinewidth{0.803000pt}%
\definecolor{currentstroke}{rgb}{0.000000,0.000000,0.000000}%
\pgfsetstrokecolor{currentstroke}%
\pgfsetdash{}{0pt}%
\pgfsys@defobject{currentmarker}{\pgfqpoint{0.000000in}{-0.048611in}}{\pgfqpoint{0.000000in}{0.000000in}}{%
\pgfpathmoveto{\pgfqpoint{0.000000in}{0.000000in}}%
\pgfpathlineto{\pgfqpoint{0.000000in}{-0.048611in}}%
\pgfusepath{stroke,fill}%
}%
\begin{pgfscope}%
\pgfsys@transformshift{2.265283in}{0.415123in}%
\pgfsys@useobject{currentmarker}{}%
\end{pgfscope}%
\end{pgfscope}%
\begin{pgfscope}%
\definecolor{textcolor}{rgb}{0.000000,0.000000,0.000000}%
\pgfsetstrokecolor{textcolor}%
\pgfsetfillcolor{textcolor}%
\pgftext[x=2.265283in,y=0.317901in,,top]{\color{textcolor}{\rmfamily\fontsize{10.000000}{12.000000}\selectfont\catcode`\^=\active\def^{\ifmmode\sp\else\^{}\fi}\catcode`\%=\active\def
\end{pgfscope}%
\begin{pgfscope}%
\pgfsetbuttcap%
\pgfsetroundjoin%
\definecolor{currentfill}{rgb}{0.000000,0.000000,0.000000}%
\pgfsetfillcolor{currentfill}%
\pgfsetlinewidth{0.803000pt}%
\definecolor{currentstroke}{rgb}{0.000000,0.000000,0.000000}%
\pgfsetstrokecolor{currentstroke}%
\pgfsetdash{}{0pt}%
\pgfsys@defobject{currentmarker}{\pgfqpoint{0.000000in}{-0.048611in}}{\pgfqpoint{0.000000in}{0.000000in}}{%
\pgfpathmoveto{\pgfqpoint{0.000000in}{0.000000in}}%
\pgfpathlineto{\pgfqpoint{0.000000in}{-0.048611in}}%
\pgfusepath{stroke,fill}%
}%
\begin{pgfscope}%
\pgfsys@transformshift{3.242816in}{0.415123in}%
\pgfsys@useobject{currentmarker}{}%
\end{pgfscope}%
\end{pgfscope}%
\begin{pgfscope}%
\definecolor{textcolor}{rgb}{0.000000,0.000000,0.000000}%
\pgfsetstrokecolor{textcolor}%
\pgfsetfillcolor{textcolor}%
\pgftext[x=3.242816in,y=0.317901in,,top]{\color{textcolor}{\rmfamily\fontsize{10.000000}{12.000000}\selectfont\catcode`\^=\active\def^{\ifmmode\sp\else\^{}\fi}\catcode`\%=\active\def
\end{pgfscope}%
\begin{pgfscope}%
\pgfsetbuttcap%
\pgfsetroundjoin%
\definecolor{currentfill}{rgb}{0.000000,0.000000,0.000000}%
\pgfsetfillcolor{currentfill}%
\pgfsetlinewidth{0.803000pt}%
\definecolor{currentstroke}{rgb}{0.000000,0.000000,0.000000}%
\pgfsetstrokecolor{currentstroke}%
\pgfsetdash{}{0pt}%
\pgfsys@defobject{currentmarker}{\pgfqpoint{0.000000in}{-0.048611in}}{\pgfqpoint{0.000000in}{0.000000in}}{%
\pgfpathmoveto{\pgfqpoint{0.000000in}{0.000000in}}%
\pgfpathlineto{\pgfqpoint{0.000000in}{-0.048611in}}%
\pgfusepath{stroke,fill}%
}%
\begin{pgfscope}%
\pgfsys@transformshift{4.220348in}{0.415123in}%
\pgfsys@useobject{currentmarker}{}%
\end{pgfscope}%
\end{pgfscope}%
\begin{pgfscope}%
\definecolor{textcolor}{rgb}{0.000000,0.000000,0.000000}%
\pgfsetstrokecolor{textcolor}%
\pgfsetfillcolor{textcolor}%
\pgftext[x=4.220348in,y=0.317901in,,top]{\color{textcolor}{\rmfamily\fontsize{10.000000}{12.000000}\selectfont\catcode`\^=\active\def^{\ifmmode\sp\else\^{}\fi}\catcode`\%=\active\def
\end{pgfscope}%
\begin{pgfscope}%
\pgfsetbuttcap%
\pgfsetroundjoin%
\definecolor{currentfill}{rgb}{0.000000,0.000000,0.000000}%
\pgfsetfillcolor{currentfill}%
\pgfsetlinewidth{0.803000pt}%
\definecolor{currentstroke}{rgb}{0.000000,0.000000,0.000000}%
\pgfsetstrokecolor{currentstroke}%
\pgfsetdash{}{0pt}%
\pgfsys@defobject{currentmarker}{\pgfqpoint{0.000000in}{-0.048611in}}{\pgfqpoint{0.000000in}{0.000000in}}{%
\pgfpathmoveto{\pgfqpoint{0.000000in}{0.000000in}}%
\pgfpathlineto{\pgfqpoint{0.000000in}{-0.048611in}}%
\pgfusepath{stroke,fill}%
}%
\begin{pgfscope}%
\pgfsys@transformshift{5.197881in}{0.415123in}%
\pgfsys@useobject{currentmarker}{}%
\end{pgfscope}%
\end{pgfscope}%
\begin{pgfscope}%
\definecolor{textcolor}{rgb}{0.000000,0.000000,0.000000}%
\pgfsetstrokecolor{textcolor}%
\pgfsetfillcolor{textcolor}%
\pgftext[x=5.197881in,y=0.317901in,,top]{\color{textcolor}{\rmfamily\fontsize{10.000000}{12.000000}\selectfont\catcode`\^=\active\def^{\ifmmode\sp\else\^{}\fi}\catcode`\%=\active\def
\end{pgfscope}%
\begin{pgfscope}%
\pgfsetbuttcap%
\pgfsetroundjoin%
\definecolor{currentfill}{rgb}{0.000000,0.000000,0.000000}%
\pgfsetfillcolor{currentfill}%
\pgfsetlinewidth{0.803000pt}%
\definecolor{currentstroke}{rgb}{0.000000,0.000000,0.000000}%
\pgfsetstrokecolor{currentstroke}%
\pgfsetdash{}{0pt}%
\pgfsys@defobject{currentmarker}{\pgfqpoint{0.000000in}{-0.048611in}}{\pgfqpoint{0.000000in}{0.000000in}}{%
\pgfpathmoveto{\pgfqpoint{0.000000in}{0.000000in}}%
\pgfpathlineto{\pgfqpoint{0.000000in}{-0.048611in}}%
\pgfusepath{stroke,fill}%
}%
\begin{pgfscope}%
\pgfsys@transformshift{6.175414in}{0.415123in}%
\pgfsys@useobject{currentmarker}{}%
\end{pgfscope}%
\end{pgfscope}%
\begin{pgfscope}%
\definecolor{textcolor}{rgb}{0.000000,0.000000,0.000000}%
\pgfsetstrokecolor{textcolor}%
\pgfsetfillcolor{textcolor}%
\pgftext[x=6.175414in,y=0.317901in,,top]{\color{textcolor}{\rmfamily\fontsize{10.000000}{12.000000}\selectfont\catcode`\^=\active\def^{\ifmmode\sp\else\^{}\fi}\catcode`\%=\active\def
\end{pgfscope}%
\begin{pgfscope}%
\definecolor{textcolor}{rgb}{0.000000,0.000000,0.000000}%
\pgfsetstrokecolor{textcolor}%
\pgfsetfillcolor{textcolor}%
\pgftext[x=3.539813in,y=0.138889in,,top]{\color{textcolor}{\rmfamily\fontsize{10.000000}{12.000000}\selectfont\catcode`\^=\active\def^{\ifmmode\sp\else\^{}\fi}\catcode`\%=\active\def
\end{pgfscope}%
\begin{pgfscope}%
\pgfsetrectcap%
\pgfsetmiterjoin%
\pgfsetlinewidth{0.803000pt}%
\definecolor{currentstroke}{rgb}{0.000000,0.000000,0.000000}%
\pgfsetstrokecolor{currentstroke}%
\pgfsetdash{}{0pt}%
\pgfpathmoveto{\pgfqpoint{0.000000in}{0.415123in}}%
\pgfpathlineto{\pgfqpoint{7.079625in}{0.415123in}}%
\pgfusepath{stroke}%
\end{pgfscope}%
\begin{pgfscope}%
\pgfsetbuttcap%
\pgfsetmiterjoin%
\definecolor{currentfill}{rgb}{1.000000,0.800000,0.800000}%
\pgfsetfillcolor{currentfill}%
\pgfsetlinewidth{1.003750pt}%
\definecolor{currentstroke}{rgb}{1.000000,0.500000,0.500000}%
\pgfsetstrokecolor{currentstroke}%
\pgfsetdash{}{0pt}%
\pgfpathmoveto{\pgfqpoint{4.681995in}{0.328111in}}%
\pgfpathlineto{\pgfqpoint{4.976170in}{0.328111in}}%
\pgfpathquadraticcurveto{\pgfqpoint{5.017837in}{0.328111in}}{\pgfqpoint{5.017837in}{0.369777in}}%
\pgfpathlineto{\pgfqpoint{5.017837in}{0.493234in}}%
\pgfpathquadraticcurveto{\pgfqpoint{5.017837in}{0.534901in}}{\pgfqpoint{4.976170in}{0.534901in}}%
\pgfpathlineto{\pgfqpoint{4.681995in}{0.534901in}}%
\pgfpathquadraticcurveto{\pgfqpoint{4.640329in}{0.534901in}}{\pgfqpoint{4.640329in}{0.493234in}}%
\pgfpathlineto{\pgfqpoint{4.640329in}{0.369777in}}%
\pgfpathquadraticcurveto{\pgfqpoint{4.640329in}{0.328111in}}{\pgfqpoint{4.681995in}{0.328111in}}%
\pgfpathlineto{\pgfqpoint{4.681995in}{0.328111in}}%
\pgfpathclose%
\pgfusepath{stroke,fill}%
\end{pgfscope}%
\begin{pgfscope}%
\definecolor{textcolor}{rgb}{0.000000,0.000000,0.000000}%
\pgfsetstrokecolor{textcolor}%
\pgfsetfillcolor{textcolor}%
\pgftext[x=4.829083in,y=0.431506in,,]{\color{textcolor}{\rmfamily\fontsize{10.000000}{12.000000}\selectfont\catcode`\^=\active\def^{\ifmmode\sp\else\^{}\fi}\catcode`\%=\active\def
\end{pgfscope}%
\begin{pgfscope}%
\pgfsetbuttcap%
\pgfsetmiterjoin%
\definecolor{currentfill}{rgb}{1.000000,0.800000,0.800000}%
\pgfsetfillcolor{currentfill}%
\pgfsetlinewidth{1.003750pt}%
\definecolor{currentstroke}{rgb}{1.000000,0.500000,0.500000}%
\pgfsetstrokecolor{currentstroke}%
\pgfsetdash{}{0pt}%
\pgfpathmoveto{\pgfqpoint{5.608186in}{0.311728in}}%
\pgfpathlineto{\pgfqpoint{5.902361in}{0.311728in}}%
\pgfpathquadraticcurveto{\pgfqpoint{5.944028in}{0.311728in}}{\pgfqpoint{5.944028in}{0.353395in}}%
\pgfpathlineto{\pgfqpoint{5.944028in}{0.476852in}}%
\pgfpathquadraticcurveto{\pgfqpoint{5.944028in}{0.518518in}}{\pgfqpoint{5.902361in}{0.518518in}}%
\pgfpathlineto{\pgfqpoint{5.608186in}{0.518518in}}%
\pgfpathquadraticcurveto{\pgfqpoint{5.566519in}{0.518518in}}{\pgfqpoint{5.566519in}{0.476852in}}%
\pgfpathlineto{\pgfqpoint{5.566519in}{0.353395in}}%
\pgfpathquadraticcurveto{\pgfqpoint{5.566519in}{0.311728in}}{\pgfqpoint{5.608186in}{0.311728in}}%
\pgfpathlineto{\pgfqpoint{5.608186in}{0.311728in}}%
\pgfpathclose%
\pgfusepath{stroke,fill}%
\end{pgfscope}%
\begin{pgfscope}%
\definecolor{textcolor}{rgb}{0.000000,0.000000,0.000000}%
\pgfsetstrokecolor{textcolor}%
\pgfsetfillcolor{textcolor}%
\pgftext[x=5.755274in,y=0.415123in,,]{\color{textcolor}{\rmfamily\fontsize{10.000000}{12.000000}\selectfont\catcode`\^=\active\def^{\ifmmode\sp\else\^{}\fi}\catcode`\%=\active\def
\end{pgfscope}%
\begin{pgfscope}%
\pgfsetbuttcap%
\pgfsetmiterjoin%
\definecolor{currentfill}{rgb}{1.000000,0.800000,0.800000}%
\pgfsetfillcolor{currentfill}%
\pgfsetlinewidth{1.003750pt}%
\definecolor{currentstroke}{rgb}{1.000000,0.500000,0.500000}%
\pgfsetstrokecolor{currentstroke}%
\pgfsetdash{}{0pt}%
\pgfpathmoveto{\pgfqpoint{2.814452in}{0.342191in}}%
\pgfpathlineto{\pgfqpoint{3.108627in}{0.342191in}}%
\pgfpathquadraticcurveto{\pgfqpoint{3.150294in}{0.342191in}}{\pgfqpoint{3.150294in}{0.383858in}}%
\pgfpathlineto{\pgfqpoint{3.150294in}{0.507315in}}%
\pgfpathquadraticcurveto{\pgfqpoint{3.150294in}{0.548981in}}{\pgfqpoint{3.108627in}{0.548981in}}%
\pgfpathlineto{\pgfqpoint{2.814452in}{0.548981in}}%
\pgfpathquadraticcurveto{\pgfqpoint{2.772786in}{0.548981in}}{\pgfqpoint{2.772786in}{0.507315in}}%
\pgfpathlineto{\pgfqpoint{2.772786in}{0.383858in}}%
\pgfpathquadraticcurveto{\pgfqpoint{2.772786in}{0.342191in}}{\pgfqpoint{2.814452in}{0.342191in}}%
\pgfpathlineto{\pgfqpoint{2.814452in}{0.342191in}}%
\pgfpathclose%
\pgfusepath{stroke,fill}%
\end{pgfscope}%
\begin{pgfscope}%
\definecolor{textcolor}{rgb}{0.000000,0.000000,0.000000}%
\pgfsetstrokecolor{textcolor}%
\pgfsetfillcolor{textcolor}%
\pgftext[x=2.961540in,y=0.445586in,,]{\color{textcolor}{\rmfamily\fontsize{10.000000}{12.000000}\selectfont\catcode`\^=\active\def^{\ifmmode\sp\else\^{}\fi}\catcode`\%=\active\def
\end{pgfscope}%
\begin{pgfscope}%
\pgfsetbuttcap%
\pgfsetmiterjoin%
\definecolor{currentfill}{rgb}{1.000000,0.800000,0.800000}%
\pgfsetfillcolor{currentfill}%
\pgfsetlinewidth{1.003750pt}%
\definecolor{currentstroke}{rgb}{1.000000,0.500000,0.500000}%
\pgfsetstrokecolor{currentstroke}%
\pgfsetdash{}{0pt}%
\pgfpathmoveto{\pgfqpoint{3.752672in}{0.347954in}}%
\pgfpathlineto{\pgfqpoint{4.046847in}{0.347954in}}%
\pgfpathquadraticcurveto{\pgfqpoint{4.088514in}{0.347954in}}{\pgfqpoint{4.088514in}{0.389620in}}%
\pgfpathlineto{\pgfqpoint{4.088514in}{0.513077in}}%
\pgfpathquadraticcurveto{\pgfqpoint{4.088514in}{0.554744in}}{\pgfqpoint{4.046847in}{0.554744in}}%
\pgfpathlineto{\pgfqpoint{3.752672in}{0.554744in}}%
\pgfpathquadraticcurveto{\pgfqpoint{3.711006in}{0.554744in}}{\pgfqpoint{3.711006in}{0.513077in}}%
\pgfpathlineto{\pgfqpoint{3.711006in}{0.389620in}}%
\pgfpathquadraticcurveto{\pgfqpoint{3.711006in}{0.347954in}}{\pgfqpoint{3.752672in}{0.347954in}}%
\pgfpathlineto{\pgfqpoint{3.752672in}{0.347954in}}%
\pgfpathclose%
\pgfusepath{stroke,fill}%
\end{pgfscope}%
\begin{pgfscope}%
\definecolor{textcolor}{rgb}{0.000000,0.000000,0.000000}%
\pgfsetstrokecolor{textcolor}%
\pgfsetfillcolor{textcolor}%
\pgftext[x=3.899760in,y=0.451349in,,]{\color{textcolor}{\rmfamily\fontsize{10.000000}{12.000000}\selectfont\catcode`\^=\active\def^{\ifmmode\sp\else\^{}\fi}\catcode`\%=\active\def
\end{pgfscope}%
\end{pgfpicture}%
\makeatother%
\endgroup%

%% file: chapters/conclusion.tex
\section{Conclusion}\label{sec:Conclusion}
We present a transformer based, calibration-robust map-view segmentation approach, that integrates RGB, LWIR and LiDAR sensors across multiple time instances. We validate our approach on field collected, maritime specific data, demonstrating remarkable potential in long-range scene understanding and suggesting that our that BEV perception has a well-functional position in the design of autonomous ship-borne navigation systems. 

%% file: chapters/future_work.tex
\section{Limitations and Future Work}\label{sec:FutureWork}

\textbf{Spatio-temporal features} Despite the positive performance boost (\cref{tab:experiment-results}) of using 3D convolution and ego-motion alignment in our suggested temporal aggregation module, our approach is still dependant on accurate ego-motion estimation information. At the same time, convolutions have local receptive fields, and are thus unable to model long spatio-temporal dependencies. Further work is required in quantifying the effect of spatio-temporal aggregation in BEV perception.

\textbf{Doppler radar:} Naturally, properties such as the radial velocity in doppler velocity radars, exhibit homogeneity for points originating from a single object. Therefore, it is reasonable to hypothesize that integrating such information will be beneficial in the BEV map segmentation task. In our experiments, the W-Radar modality was excluded due to its sparse sampling, which made it unsuitable for use. However, as recently highlighted by \cite{hendy2020fishing,harley2023simple}, BEV perception could benefit from further investigation into the fusion of the W-Radar modality, suggesting it as a promising area for future research. It is worth mentioning that our approach readily supports the integration of W-Radar data as an additional modality, by aggregating them into additional pseudo-views. 

\textbf{Cross-attention:} While capable of capturing global correspondences across different map locations and views, the complexity of cross attention scales linearly with the number resolution of the BEV map, and quadratically with input feature resolution and number of timesteps, and number of views, prohibiting the use of fine-detailed feature maps or BEV maps. At the same time, a lot of calculations are possibly redundant as global cross attention, naively calculates scores between BEV map locations and views, that do not necessarily correspond to each other. The use of deformable attention \cite{Zhu2020DeformableDetection}, as already demonstrated in \cite{li2022bevformer,yang2023bevformer,harley2023simple} is a promising direction in reducing the computation complexity of cross attention, and at the same time enables the selective attendance of a BEV query to specific views regions.